\documentclass[11pt,a4paper]{article}
\usepackage[hyperref]{emnlp2020}
\usepackage{times}
\usepackage{latexsym}

\usepackage{microtype}

\aclfinalcopy 

\usepackage{booktabs}
\usepackage{pifont}
\usepackage{url}
\usepackage{amsmath}
\usepackage{amssymb}
\usepackage{multirow}
\usepackage{xspace}
\usepackage{soul}
\usepackage{epsfig,endnotes,graphicx,xspace}
\usepackage{enumitem}
\usepackage{hyperref}
\usepackage{float}
\usepackage{framed}
\usepackage{fdsymbol}
\usepackage{subfig}

\newcommand{\eat}[1]{}
\newcommand{\Mh}{Multi-hop\xspace}
\newcommand{\mh}{multi-hop\xspace}

\newcommand{\mf}{multifact\xspace}
\newcommand{\MfR}{Multifact Reasoning\xspace}

\newcommand{\mfr}{multifact reasoning\xspace}

\newcommand{\DiscReas}{Disconnected Reasoning\xspace}
\newcommand{\discreas}{disconnected reasoning\xspace}
\newcommand{\direname}{\underline{Di}sconnected \underline{Re}asoning\xspace}

\newcommand{\CSSTAbb}{CSST\xspace}

\newcommand*{\img}[1]{%
    \raisebox{-.1\baselineskip}{%
        \includegraphics[
        height=0.7\baselineskip,
        width=0.7\baselineskip,
        keepaspectratio,
        ]{#1}\,%
    }%
}

\newcommand{\D}{\mathcal{D}}
\newcommand{\T}{\mathbb{T}}
\newcommand{\TD}{\mathbb{T}(\mathcal{D})}

\newcommand{\probe}{DiRe\xspace}
\newcommand{\direprobe}{DiRe probe\xspace}

\newcommand{\Lans}{L_\textrm{ans}}
\newcommand{\Lansifpresent}{L^?_\textrm{ans}}
\newcommand{\Lsupp}{L_\textrm{supp}}
\newcommand{\Lsuff}{L_\textrm{suff}}
\newcommand{\Lsuffprb}{L^*_\textrm{suff}}

\newcommand{\test}{\tau}
\newcommand{\prediction}{\mu}

\newcommand{\mtest}{m_\test}
\newcommand{\mans}{m_\textrm{ans}}
\newcommand{\msupp}{m_\textrm{supp}}

\newcommand{\PRB}{\mathbb{P}}
\newcommand{\PRBtest}{\mathbb{P}_\test}

\newcommand{\PRBanssupp}{\mathbb{P}_{\textrm{ans}+\textrm{supp}}}
\newcommand{\PRBtestsuff}{\mathbb{P}_{\test+\textrm{suff}}}
\newcommand{\PRBall}{\mathbb{P}_{\textrm{ans}+\textrm{supp}+\textrm{suff}}}
\newcommand{\PRBD}{\mathbb{P}(\mathcal{D}) }
\newcommand{\PRBTD}{\mathbb{P}(\T(\mathcal{D})) }

\newcommand{\mtestp}{m_\test^{\PRB}}
\newcommand{\mtestt}{m_{\mathrm{\test+suff}}^{\T}}
\newcommand{\mtestpt}{m_{\mathrm{\test+suff}}^{\PRB\T}}

\newcommand{\Tadv}{\mathbb{T}\textsubscript{adv}}

\newcommand{\TadvD}{\mathbb{T}\textsubscript{adv}(\mathcal{D}) }

\newcommand{\TTadvD}{\mathbb{T}(\mathbb{T}\textsubscript{adv}(\mathcal{D})) }

\newcommand{\redfact}{\img{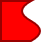}}
\newcommand{\bluefact}{\img{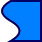}}
\newcommand{\lightbluefact}{\img{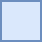}}

\newcommand{\yellowfact}{\img{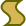}}
\newcommand{\purplefact}{\img{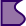}}

\newcommand{\hpqa}{HotpotQA\xspace}

\newcommand{\ans}{Ans\xspace}
\newcommand{\ssp}{Supp\textsubscript{s}\xspace}
\newcommand{\psp}{Supp\textsubscript{p}\xspace}
\newcommand{\psuff}{Suff\xspace}
\newcommand{\sponly}{Supp\xspace}

\newcommand{\direcondition}{\textsc{\underline{DiRe}} condition\xspace}

\newcommand{\added}[1]{#1}

\title{Is Multihop QA in \textsc{\underline{DiRe}} Condition?\\ Measuring and Reducing Disconnected Reasoning}

\author{
  Harsh Trivedi$^\dagger$\thanks{\added{\enskip Early portion of this work was done during the first author's internship at Allen Institute for AI.}}\ \ \ \
  Niranjan Balasubramanian$^\dagger$\ \ \ \
  Tushar Khot$^\ddagger$\ \ \ \
  Ashish Sabharwal$^\ddagger$\\
  \\
  $^\dagger$ Stony Brook University, Stony Brook, U.S.A.\\
  \texttt{\small \{hjtrivedi,niranjan\}@cs.stonybrook.edu}\\
  $^\ddagger$ Allen Institute for AI, Seattle, U.S.A.\\
  \texttt{\small \{tushark,ashishs\}@allenai.org}
}

\date{}

\begin{document}

\addtolength{\abovedisplayskip}{-3pt}
\addtolength{\belowdisplayskip}{-3pt}

\maketitle

\begin{abstract}
Has there been real progress in \mh question-answering? Models often exploit dataset artifacts to produce correct answers, without connecting information across multiple supporting facts. This limits our ability to measure true progress and defeats the purpose of building \mh QA datasets. We make three contributions towards addressing this. First, we formalize such undesirable behavior as disconnected reasoning across subsets of supporting facts. This allows developing a model-agnostic probe for measuring how much any model can cheat via disconnected reasoning. Second, using a notion of \emph{contrastive support sufficiency}, we introduce an automatic transformation of existing datasets that reduces the amount of disconnected reasoning. \added{Third, our experiments\footnote{\url{https://github.com/stonybrooknlp/dire}} suggest that there hasn't been much progress in \mf QA in the reading comprehension setting.} For a recent large-scale model (XLNet), we show that only 18 \added{points out} of its answer \added{F1 score of 72} on \hpqa are obtained through \mfr, roughly the same as that of a simpler RNN baseline. Our transformation substantially reduces disconnected reasoning (19 points in answer F1). It is complementary to adversarial approaches, yielding further reductions in conjunction.
\end{abstract}

\section{Introduction}

\begin{figure}[t]
    \centering
	\includegraphics[width=0.65\textwidth]{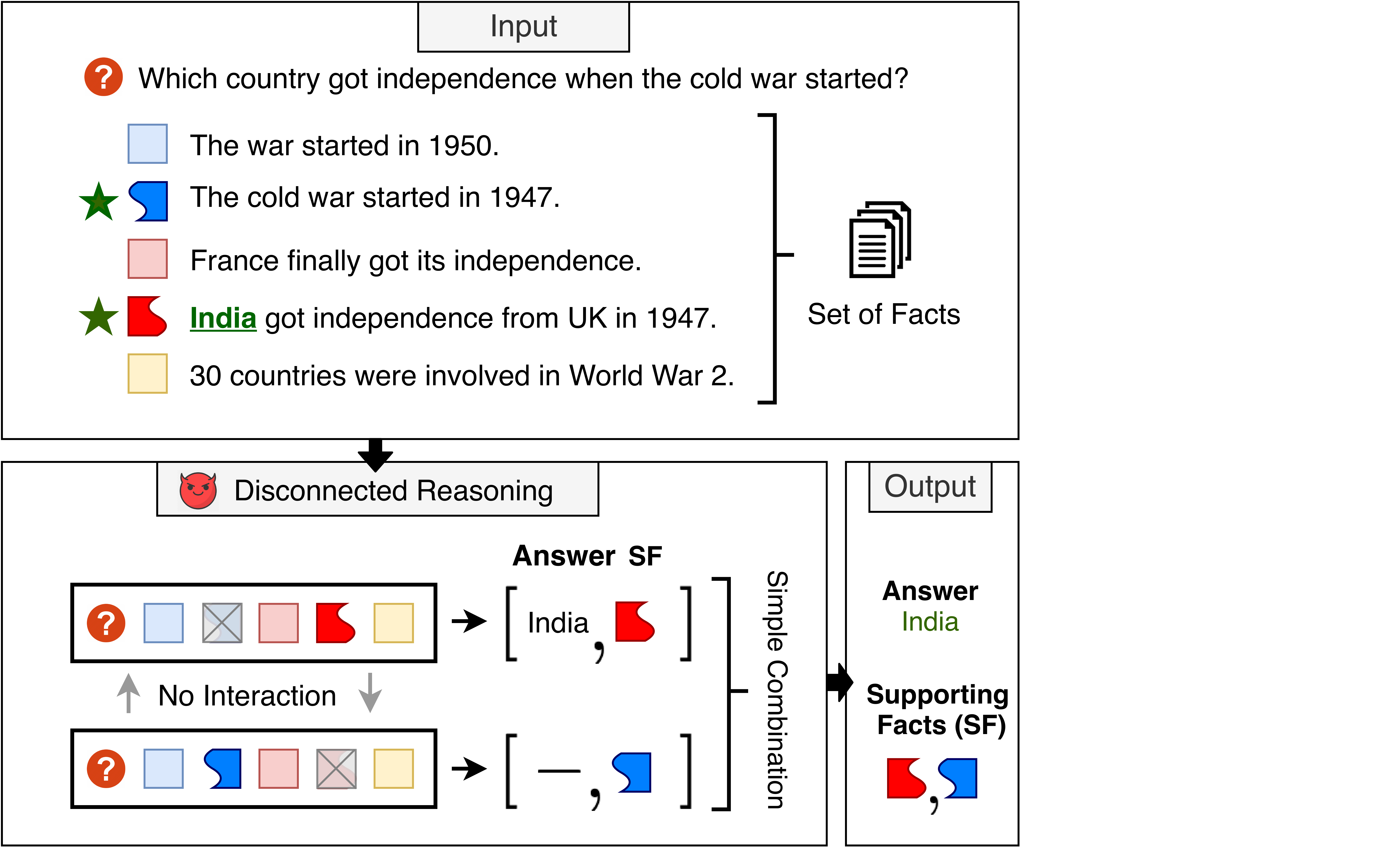}
	\caption{Example of \emph{\discreas}, a form of bad \mfr: 
     Model arrives at the answer by simply combining its outputs from two subsets of the input, neither of which contains all supporting facts. From one subset, it identifies the blue supporting fact (\bluefact), the only one mentioning cold war. \textit{Independently}, from the other subset, it finds the red fact (\redfact) as the only one mentioning a country getting independence with associated time, and returns the correct answer (India). Further, it returns a simple union of the supporting facts it found over the input subsets.}
	\vspace{-4pt}
	\label{fig:intro}
\end{figure}

\Mh question answering requires connecting and synthesizing information from multiple facts in the input text, a process we refer to as \emph{\mfr}. Prior work has, however, shown that bad reasoning models, ones that by design do not connect information from multiple facts, can achieve high scores because they can exploit specific types of biases and artifacts (e.g., answer type shortucts) in existing datasets~\cite{min2019compositional,chen2019understanding}. While this demonstrates the existence of models that \emph{can} cheat, what we do not know is the extent to which current models \emph{do} cheat, and whether there has been real progress in building models for \mfr.

We address this issue in the context of multi-hop reading comprehension. We introduce a general-purpose characterization of a form of bad multihop reasoning, namely \emph{disconnected reasoning}. For datasets annotated with supporting facts, this allows devising a model-agnostic probe to estimate the extent of disconnected reasoning done by \emph{any} model, and an automatic transformation of existing datasets that reduces such \discreas.

\paragraph{\bf Measuring \DiscReas.}

Good \mfr,\footnote{\added{We refer to desirable types of \mfr as \emph{good} and undesirable types as \emph{bad}.}} at a minimum, requires models to connect information from one or more
facts when they select and use information from other facts to arrive at an answer.
However, models can cheat, as illustrated in Figure~\ref{fig:intro}, by independently assessing information in subsets of the input facts none of which contains all supporting facts, and taking a simple combination of outputs from these subsets (e.g., by taking a union) to produce the overall output. This entirely avoids
meaningfully combining information across all supporting facts, a fundamental requirement of \mfr. 
We refer to this type of reasoning as \emph{disconnected reasoning} (\probe\ in short) and provide a formal criterion, the \direcondition, to catch cheating models.
Informally, it checks whether for a given test of \mfr (e.g., answer prediction or supporting fact identification), a model is able to trivially combine its outputs on subsets of the input context (none of which has all supporting facts) without any interaction between them. 

Using the \direcondition, we develop a systematic probe, involving an automatically generated probing dataset, that measures how much a model can score using \discreas.


\paragraph{Reducing \DiscReas.}

A key aspect of a \discreas model is that it does not change its behavior towards the selection and use of supporting facts that are in the input, whether or not the input contains all of the supporting facts the question requires. This suggests that the notion of sufficiency---whether all supporting facts are present in the input, which clearly matters to a good \mf model---does not matter to a bad model. We formalize this into a \emph{constrastive support sufficiency} test (\CSSTAbb) as an additional test of \mfr that is harder to cheat. We introduce an automatic transformation that adds to each question in an original \mh dataset a group of \emph{insufficient context} instances corresponding to different subsets of supporting facts. A model must recognize these as having insufficient context in order to receive any credit for the question.

Our \textbf{empirical evaluation} on the \hpqa dataset~\cite{hotpotqa} reveals three interesting findings: 
(i) A substantial amount of progress on \mh reading comprehension can be attributed to improvements in \discreas. E.g., XLNet~\cite{yang2019xlnetga}, a recent large-scale langugage model, only achieves 17.5 F1 pts (of its total 71.9 answer F1) via \mfr, roughly the same as a much simpler RNN model.
(ii) Training on the transformed dataset with \CSSTAbb results in a substantial reduction in \discreas (e.g., a 19 point drop in answer F1),
demonstrating that it less cheatable, is a harder test of \mfr, and gives a better picture of the current state of \mfr.
(iii) The transformed dataset is more effective at reducing \discreas than a previous adversarial augmentation method~\cite{jiang2019avoiding}, and is also complementary, improving further in combination.

\paragraph{}
\added{In summary, the \direprobe serves as a simple yet effective tool for \textbf{model designers} to assess how much of their model's score can actually be attributed to \mf reasoning. Similarly, \textbf{dataset designers} can assess how cheatable is their dataset $D$ (in terms of allowing \discreas) by training a strong model on the \direprobe for $D$, and use our transform to reduce $D$'s cheatability.}

\section{Related Work}

\textbf{Multi-hop Reasoning:}
Many \mf reasoning approaches have been proposed for \hpqa and similar datasets~\cite{obqa,qasc}. These use iterative fact selection~\cite{qfe,sae,asai2020learning,das2019multi}, graph neural networks~\cite{dfgn,hgn,sae}, or simply cross-document self-attention~\cite{yang2019xlnetga,Beltagy2020Longformer} to capture inter-paragraph interaction. While these approaches have pushed the state of the art, \added{the extent of actual progress on \mf reasoning remains unclear.}

\textbf{Identifying Dataset Artifacts:}
Several works have identified dataset artifacts for tasks such as NLI~\cite{gururangan2018annotation}, Reading Comprehension~\cite{feng2018pathologies,sugawara2019assessing}, and even multi-hop reasoning~\cite{min2019compositional,chen2019understanding}. These artifacts allow models to solve the dataset without actually solving the underlying task. 
On HotpotQA, prior work has shown \emph{existence} of models that identify the support~\cite{quark} and  answer~\cite{min2019compositional,chen2019understanding} by operating on each paragraph or sentence independently. We, on the other hand, estimate the amount of \discreas in \emph{any} model and quantify the cheatability of answer and support identification.

\textbf{Mitigation of Dataset Artifacts:}
To deal with these artifacts, several adversarial methods have been proposed for reading comprehension~\cite{jia2017adversarial, rajpurkar2018know} and multi-hop QA~\cite{jiang2019avoiding}. These methods  minimally perturb the input text to limit the effectiveness of the dataset artifacts. Our insufficient context instances that partition the context are complementary to these approaches (as we show in our experiments). \citet{rajpurkar2018know}  used a mix of answerable and unanswerable questions to make the models avoid superficial reasoning. In a way, while these hand-authored unanswerable questions also provide insufficient context, we specifically focus on (automatically) creating unanswerable \emph{multi-hop questions} by providing insufficient context.

\textbf{Minimal Pairs:}
Recent works~\cite{kaushik2019learning,ropes,Gardner2020EvaluatingNM} have proposed evaluating NLP systems by generating minimal pairs (or contrastive examples) that are similar but have different labels. Insufficient context instances in our sufficiency test can be thought of as automatically generated contrastive examples specifically for avoiding disconnected reasoning.

%



\section{Measuring \DiscReas}
\label{sec:measuring}

\added{This section formalizes the \emph{\direcondition}, which captures what it means for a model to employ \discreas, and describes how to use this condition to probe the amount of \discreas performed by a given model, and the extent of such reasoning possible on a dataset.}

A good \mfr is one where information from all the supporting facts is meaningfully synthesized to arrive at an answer. The precise definition for what constitutes meaningful synthesis is somewhat subjective; it depends on the semantics of the facts and the specific question at hand, making it challenging to devise a measurable test for the amount of \mf (or non-multifact) reasoning done by a model or needed by a dataset.

\added{Previous works have used the Answer Prediction task (i.e., identifying the correct answer) and the Supporting Fact Identification task (identifying all facts supporting the answer) as approximate \emph{tests} of \mfr}.
We argue that, {\em at a minimum}, good \mfr requires connected reasoning---one where information from at least one supporting fact is connected to the selection and use of information from other supporting facts. Consider the example question in Figure~\ref{fig:intro}. A good \mfr will look for a supporting fact that mentions when the cold war started (\bluefact) and use information from this fact (year 1947) to select the other supporting fact mentioning the country that got independence (\redfact) (or vice versa). 

A bad \mfr model, however, can cheat on answer prediction by only looking for a fact that mentions a country getting independence at some time (mentioned in \redfact), without connecting this to when the cold war started (mentioned in \bluefact).
Similarly, the model can also cheat on supporting fact identification by treating it as two independent sub-tasks---one returning a fact mentioning the time when a country got independence, and another for a fact mentioning the time when cold war started. The result of \added{at least} one of the two sub-tasks should influence the result of the other sub-task, but here it does not. This results in \emph{disconnected reasoning}, where \emph{both} supporting facts are identified without reference to the other.\footnote{Identifying one of the facts in isolation is fine, as long as information from this fact is used to identify the other fact.} Even though the precise definition of a meaningful synthesis of information is unclear, it is clear that models performing this type of disconnected reasoning cannot be considered as doing valid \mfr. 
Neither answer prediction nor support identification directly checks for such disconnected reasoning.

\subsection{Disconnected Reasoning}
\label{subsec:dire-characterize}

We can formalize the notion of disconnected reasoning from the perspective of any multihop reasoning test. For the rest of this work, we assume a multifact reading comprehension setting, where we have a dataset $D$ with instances of the form $q = (Q,C; A)$. Given a question $Q$ along with a context $C$ consisting of a set of facts, the task is to predict the answer $A$. $C$ includes a subset $F_s$ of at least two facts that together provide support for $A$.

Let $\test$ denote a test of \mfr and $\test(q)$ the output a model should produce when tested on input $q$. Consider the Support Identification test, where $\test(q) = F_s$. Let there be two proper subsets of the supporting facts $F_{s1}$ and $F_{s2}$ such that $F_s = F_{s1} \cup F_{s2}$. We argued above that a model performs disconnected reasoning if it does not use information in $F_{s1}$ to select and use information in $F_{s2}$ and vice versa. One way we can catch this behavior is by checking if the model is able to identify $F_{s1}$ given $C \setminus F_{s2}$ and identify $F_{s2}$ given $C \setminus F_{s1}$. To pass the test successfully the model only needs to trivially combine its outputs from the two subsets $C \setminus F_{s2}$ and $C \setminus F_{s1}$. 

Concretely, we say $M$ performs \direname on $q$ from the perspective of a test $\test$ if the following condition holds:
\FrameSep4pt
\begin{framed}
    \noindent
    \textbf{\direcondition}: There exists a proper bi-partition\footnotemark $\{F_{s1}, F_{s2}\}$ of $F_s$ such that the two outputs of $M$ with input $q$ modified to have $C \setminus F_{s2}$ and $C \setminus F_{s1}$ as contexts, respectively, can be \emph{trivially combined} to produce $\test(q)$.
\end{framed}
\footnotetext{$\{X, Y\}$ is a proper bi-partition of a set $Z$ if $X \cup Y = Z, X \cap Y = \phi, X \neq \phi,$ and $Y \neq \phi$.}

The need for considering all proper bi-partitions is further explained in Appendix~\ref{app:all-bipartitions}, using an example of 3-hop reasoning (Figure~\ref{fig:dire-generalization}). The \direcondition does not explicitly state what constitutes a trivial combination; this is defined below individually for each test.
We note that it only captures disconnected reasoning, which is one manifestation of the lack of a meaningful synthesis of facts.

For Answer Prediction, trivial combination corresponds to producing answers (which we assume are associated confidence scores) independently for the two contexts, and choosing the answer with the highest score. Suppose $M$ answers $a_1$ with score $s(a_1)$ on the first context in the \direcondition; similarly $a_2$ for the second context. We say the condition is met if $A = \arg \max_{a \in \{a_1, a_2\}} s(a)$.

For the Support Identification test, as in the example discussed earlier, \emph{set union} constitutes an effective trivial combination. Suppose $M$ identifies $G_1$ and $G_2$ as the sets of supporting facts for the two inputs in the \direcondition, respectively. We say the condition is met if $G_1 \cup G_2 = F_s$.

\added{In the above discussion, we assumed the so called `exact match' or EM metric for assessing whether the answer or supporting facts produced by the combination operator were correct. In general, let $\mtest(q, \prediction(q))$ be any metric for scoring the output $\prediction(q)$ of a model against the true label $\test(q)$ for a test $\test$ on question $q$ (e.g., answer EM, support F1, etc.). We can apply the same metric to the output of the combination operator (instead of  $\prediction(q)$) to assess the extent to which the \direcondition is met for $q$ under the metric $\mtest$.}

\subsection{Probing \DiscReas}
\label{subsec:dire-probe}

The \direcondition allows devising a probe for measuring how much can a model cheat on a test $\test$, i.e., how much can it score using \discreas. The probe for a dataset $D$ comprises an \emph{automatically generated}  dataset $\PRBtest(D)$, on which the model is evaluated, \added{with or without training}.

For simplicity, consider the case where $F_s = \{f_1, f_2\}$. Here $\{\{f_1\}, \{f_2\}\}$ is the unique proper bi-partition of $F_s$. The \direcondition checks whether a model $M$ can arrive at the correct test output $\test(q)$ for input $q = (Q,C; A)$ by trivially combining its outputs on contexts $C \setminus \{f_1\}$ and $C \setminus \{f_2\}$. Accordingly, for each $q \in D$, the probing dataset $\PRBanssupp(D)$ for Answer Prediction and Support Identification contains a \emph{group of instances}: 
\begin{eqnarray}
    (Q, C \setminus \{f_1\}; \Lansifpresent{=}A, \Lsupp{=}\{f_2\}) \label{eqn:2fact-probe-1}\\
    (Q, C \setminus \{f_2\}; \Lansifpresent{=}A, \Lsupp{=}\{f_1\})
    \label{eqn:2fact-probe-2}
\end{eqnarray}
where $\Lsupp$ denotes the support identification
label and $\Lansifpresent{=}A$
represents an optional answer label that is included only if $A$ is present in the supporting facts retained in the context. \added{These labels are only used if the model is trained on $\PRBtest(D)$.}

Models operate independently over instances in $\PRBtest(D)$, whether or not they belong to a group. Probe performance, however, is measured via a \textbf{grouped probe metric}, \added{denoted $\mtestp$,} that captures \added{how well does the} \emph{trivial combination} of the two corresponding outputs \added{match} $\test(q)$ \added{according to metric $m_\tau$}, as per the \direcondition for $\test$. Specifically, for Answer Prediction, we use the highest scoring answer (following the \emph{argmax} operator in Section~\ref{subsec:dire-characterize}) across the two instances in the group, and evaluate it against $A$ using a standard metric \added{$\mans$} (EM, F1, etc.). For Support Identification, we take the \emph{union} of the two sets of supporting facts identified (for the two instances), and evaluate it against $\{f_1, f_2\}$ using a standard metric \added{$\msupp$}.

\paragraph{General case of $|F_s| \geq 2$:}

We translate each $q \in D$ into a \emph{collection} $\PRBtest(q)$ of $2^{|F_s|-1} - 1$ groups of instances, with $\PRBtest(q; s_1)$ denoting the group for the bi-partition $\{F_{s1}, F_{s2}\}$ of $F_s$.

This group, for answer and support prediction tests, contains:
\begin{eqnarray}
    (Q, C \setminus F_{s1}; \Lansifpresent{=}A, \Lsupp{=}F_{s2})\\
    (Q, C \setminus F_{s2}; \Lansifpresent{=}A, \Lsupp{=}F_{s1})
\end{eqnarray}

As per the \direcondition, as long as the model cheats on any one bi-partition, it is considered to cheat on the test. Accordingly, the probe metric $\mtestp$ uses a \emph{disjunction} over the groups:

\begin{equation}
\label{eq:probe-metric}
\resizebox{0.89\hsize}{!}{%
    $\mtestp(q, \prediction(\PRBtest(q))) = \max_{\{F_{s1}, \_\}} \mtestp(q, \prediction(\PRBtest(q; s_1)))$
    }
\end{equation}
where \added{$\prediction(\PRBtest(q))$ denotes the model's prediction on the probe group $\PRBtest(q)$,} the max is over all proper bi-partitions of $F_s$, and \added{$\mtestp(q, \prediction(\PRBtest(q; s_1)))$} denotes the probe metric for the group \added{$\PRBtest(q; s_1)$} which incorporates the trivial combination operator, denoted $\oplus$, associated with $\test$ as follows:
\begin{align}
\label{eq:probe-metric-with-operator}
     \resizebox{0.89\hsize}{!}{%
    $\mtestp(q, \prediction(\PRBtest(q; s_1))) = \mtest \big(q, \oplus_{q' \in \PRB(q; s_1)} \prediction(q') \big)$
    }
\end{align}
For example, when $\test$ is answer prediction, we view \added{$\prediction(q')$} as both the predicted answer and its score for $q'$, $\oplus$ chooses the answer with the highest score, and $\mans$ evaluates it against $A$. When $\test$ is support identification, \added{$\prediction(q')$} is the set of facts the model outputs for $q'$, $\oplus$ is union, and $\msupp$ is a standard evaluation of the result against the true label $F_s$.

\subsection{Use Cases of \probe Probe}
\label{subsec:measuring-use-cases}

The probing dataset $\PRBtest(D)$ can be used by \textbf{model designers} to assess what portion of their model $M$'s performance on $D$ can be achieved on a test $\test$ via disconnected reasoning, by computing:
\begin{align}
    \text{DiRe}^\test(M, D) = \mathrm{S^\test_{cond}}(M, \PRBtest(D) \mid D)
    \label{eqn:original-model-cheatability}
\end{align}
This is a zero-shot evaluation\footnote{Our experiments also include inoculation (i.e., finetuning on a small fraction of the dataset) before evaluation.} where $M$ is not trained on $\PRBtest(D)$. $\mathrm{S^\test_{cond}}$ represents $M$'s score on $\PRBtest(D)$ conditioned on its score on $D$, computed as the question-wise minimum of $M$'s score on $D$ and $\PRBtest(D)$, in terms of metrics $\mtest$ \added{and $\mtestp$, resp.}\footnote{For answer prediction with exact-match, this corresponds to $M$ getting 1 point for correctly answering a question group in $\PRBD$ only if it correctly answers the corresponding original question in $\D$ as well.}

Similarly, $\PRBtest(D)$ can be used by a \textbf{dataset designer} to assess how cheatable $D$ is by computing:
\begin{align}
    \text{DiRe}^\test(D) = \mathrm{S}^\test(M^*, \PRBtest(D))
    \label{eqn:original-data-cheatability}
\end{align}
where $M^*$ is the strongest available model architecture for $\test$ that is trained on $\PRBtest(D)$ and $\mathrm{S}^\test$ is its score under metric \added{$\mtestp$}.

\section{Reducing Disconnected Reasoning}
\label{sec:css}

\added{This section introduces an automatic transformation of a dataset to make it less cheatable by disconnected reasoning. It also defines a probe dataset for assessing how cheatable the transformed dataset is.}

\begin{figure}[t]
    \centering
	\includegraphics[width=0.45\textwidth]{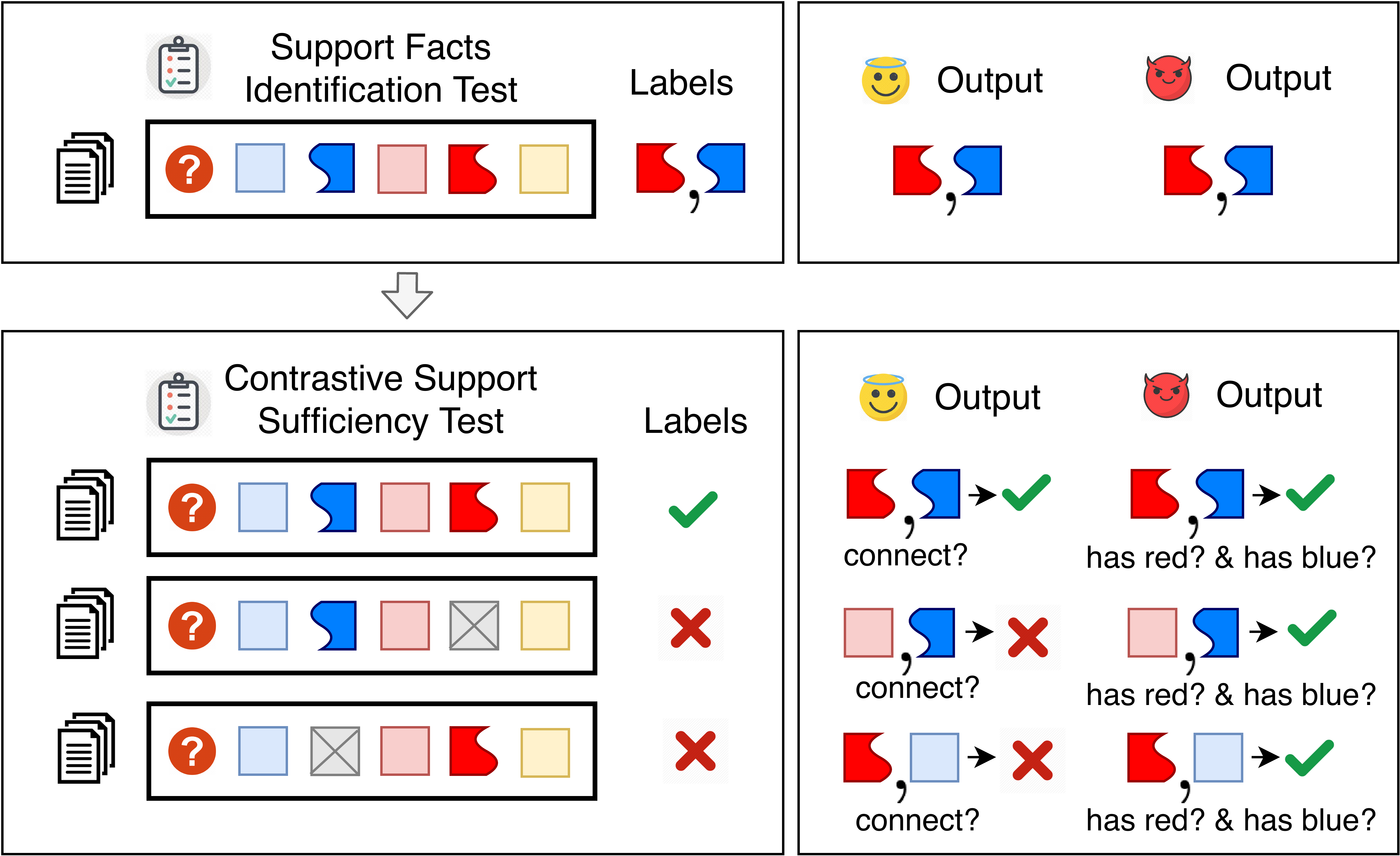}
	\caption{Transformation of a question for Contrastive Support Sufficiency evaluation. Top-Left: Original instance $q$ labeled with red (\redfact) and blue (\bluefact) supporting facts $F_s$. Bottom-Left: Its transformation into a group $\T(q)$ of 3 instances, one with sufficient and two with insufficient context, with labels denoting context sufficiency. Right: Behavior of good vs.\ bad models on $q$ and $\T(q)$. A good \mf model would realize that the potentially relevant facts are not sufficient (do not connect) whereas a bad model would find potentially relevant facts and assume they are sufficient.}
	\label{fig:sufficiency}
	\vspace{-8pt}
\end{figure}

A disconnected reasoning model does not connect information across supporting facts. This has an important consequence: when a supporting fact is dropped from the context, the model's behavior on other supporting facts remains unchanged. Figure~\ref{fig:sufficiency} illustrates this for the example in Figure~\ref{fig:intro}.

Suppose we create an insufficient context $C'$ by removing the blue fact (\bluefact) from $C$ (shown in the last row of Figure~\ref{fig:sufficiency}, with the removed fact crossed out). With the full context $C$, the cheating model discussed earlier did not use the information in the blue fact (\bluefact) to produce the answer or to identify the red fact (\redfact). Therefore, the absence of the blue fact in $C'$ will induce \emph{no change} in this model's answer or its ability to select the red fact (\redfact). Further, to return a second supporting fact, the model would choose the next best matching fact (\lightbluefact) that also indicates the start of a war and thus appears to be a reasonable choice (see bottom-right of Figure~\ref{fig:sufficiency}). Without considering interaction between the two identified facts, this model would not realize that the light blue fact (\lightbluefact) does not fit well with the red fact (\redfact) because of the year mismatch (1950 vs.\ 1947), and the two together are thus insufficient to answer $Q$. 

A good \mf model, on the other hand, connects information across different supporting facts. Thus, when evaluated on context $C'$ with the blue fact (\bluefact) missing, its answer as well as behavior for selecting the other supporting facts \emph{will} be affected.

\subsection{Contrastive Support Sufficiency}

The above example illustrates that \emph{sufficiency of supporting facts} in the input context matters to a good \mf model (i.e., it behaves differently under $C$ and $C'$) but not to a disconnected reasoning model. This suggests that if we force models to pay attention to sufficiency, we can reduce disconnected reasoning. We formalize this idea and introduce the notion of \emph{contrastive support sufficiency}. Informally, for each question, we consider several variants of the context that are contrastive: some contain sufficient information (i.e., $F_s \subseteq C$) while others don't. Evaluating models with these contrastive inputs allows discerning the difference in behavior between good and bad models. Figure~\ref{fig:sufficiency} illustrates an example of contrastive contexts and the expected behavior of such models.

\subsection{Transforming Existing Datasets}
\label{subsec:transformation}

To operationalize this idea, we introduce an \emph{automated} dataset transformation, Contrastive Support Sufficiency Transform ($\T$), applicable to any multifact reading comprehension dataset 
where each question is associated with a set of facts as context, of which a subset is annotated as supporting facts.
Intuitively, given a context $C$, we want the model to identify whether $C$ is sufficient to answer the question. If sufficient, we also want it to provide correct outputs for other tests (e.g., answer prediction).

Formally, $\T(D)$ transforms each instance $q = (Q,C; A)$ in a dataset $D$ into a \emph{group} $\T(q)$ of two types of instances, those with sufficient support and those without. For simplicity, consider the case of $F_s = \{f_1, f_2\}$ as in Section~\ref{subsec:dire-probe}. The transformed instance group $\T(q)$ is illustrated in the bottom half of Figure~\ref{fig:sufficiency}. It includes two \emph{insufficient context instances} corresponding to the two non-empty proper subsets of $F_s$, with the output label set to $\Lsuff = 0$ (illustrated as $\color{red} \times$ in Figure~\ref{fig:sufficiency}):
%
\begin{align*}
    (Q, C \setminus \{f_1\};\,  \Lsuff{=}0), (Q, C \setminus \{f_2\};\,  \Lsuff{=}0)
\end{align*}
Since these contexts lack sufficient information, we omit labels for answer or supporting facts.

$\T(q)$ also includes a single \emph{sufficient context instance}, but not with entire $C$ as the context. To avoid introducing a context length bias relative to the above two instances, we remove from $C$ a fixed, uniformly sampled non-supporting fact $f_r$ chosen from $C \setminus F_s$ (we assume $|C| \geq 3$). The output label is set to $\Lsuff = 1$. Since the context is sufficient, the correct answer and supporting facts are included as additional labels, to use for Answer and Support tests if desired, resulting in the instance:
\begin{equation}
    (Q, C \setminus \{f_r\};\, \Lans{=}A, \Lsupp{=}F_s, \Lsuff{=}1)
\end{equation}

For any performance metric \added{$\mtest(q, \cdot)$} of interest in $D$ (e.g., answer EM, support F1, etc.), the corresponding \textbf{transformed metric} \added{$\mtestt(q, \cdot)$} operates in a conditional fashion: it equals $0$ if any $\Lsuff$ label in the group is predicted incorrectly, and equals \added{$\mtest(q_\textrm{suff}, \cdot)$} otherwise, where $q_\textrm{suff}$ denotes the unique sufficient context instance in $\T(q)$. \added{A model that predicts all instances to be insufficient (or sufficient) will get 0 pts under $\mtestt$.}

The case of $|F_s| \geq 2$ is left to Appendix~\ref{app:transform-general-k}. Intuitively, \added{$\mtestt(q, \cdot) \neq 0$} suggests that
when reasoning with any proper subset of $F_s$, the model relies on at least one supporting fact outside of that subset.
High performance on $\T(D)$ thus suggests combining information from all facts.\footnote{We say \emph{suggests} rather than guarantees because the behavior of the model with partial context $C' \subset C$ may not be qualitatively identical to its behavior with full context $C$.}$^,$\footnote{\label{footnote:no-semantically-meaningful-check}The transformation encourages a model to combine information from all facts in $F_s$. Whether the information that is combined is semantically meaningful or how it is combined is interesting is beyond its scope.}

\subsection{Probing \DiscReas in $\T(D)$}
\label{subsec:transformation-probe}

The sufficiency test (\CSSTAbb) used in the transform discourages \discreas by encouraging models to track sufficiency. Much like other tests of \mfr, we can apply the \direcondition to probe models for how much they can cheat on \CSSTAbb.
As explained in Appendix~\ref{app:probe-T}, the probe checks whether a model $M$ can independently predict whether $F_{s1}$ and $F_{s2}$ are present in the input context, without relying on each other. If so, $M$ can use \discreas to correctly predict sufficiency labels in $\T(D)$. 

For $F_s = \{f_1, f_2\}$, if the transformed group $\T(q)$ uses fact $f_r$ for context length normalization, the probing group $\PRB(\T(q))$ contains 3 instances:

\begin{align*}
    (Q, C \setminus \{f_1, f_r\};\, & \Lansifpresent{=}A, \Lsupp{=}\{f_2\}, \Lsuffprb{=}0)\\
    (Q, C \setminus \{f_2, f_r\};\, & \Lansifpresent{=}A, \Lsupp{=}\{f_1\}, \Lsuffprb{=}0)\\
    (Q, C \setminus \{f_1, f_2\};\, & \Lsuffprb{=}-1)
\end{align*}

Metric \added{$\mtestpt(q, \cdot)$} on this probing group equals $0$ if the model predicts any of the $\Lsuffprb$ labels incorrectly. Otherwise, we use the grouped probe metric \added{$\mtestp(q, \cdot)$} from Section~\ref{subsec:dire-probe} for the first two instances, ignoring their $\Lsuffprb$ label. Details  are deferred to Appendices~\ref{app:probe-T} and~\ref{app:probe-T-general-k}

\added{We can use $\PRBTD$ to assess how cheatable $\TD$ is via disconnected reasoning by computing:
\begin{align}
\resizebox{0.86\hsize}{!}{
    $\mathrm{DiRe^{\tau+suff}}(\TD) = \mathrm{S^{\tau+suff}}(M'^*, \PRB(\TD))$
    }
    \label{eqn:transformed-data-cheatability}
\end{align}
where $M'^*$ is the strongest available model architecture for $\mathrm{\tau+suff}$ that is trained on $\PRB(\TD)$, and $\mathrm{S^{\tau+suff}}$ is its score under the metric \added{$\mtestpt$.} }

\begin{table*}[t]
    \centering
    \footnotesize
    \setlength\tabcolsep{3pt}
    \begin{tabular}{lp{2.0cm}p{5.1cm}p{4.2cm}p{2.3cm}}
        \textbf{Dataset}       &  \textbf{Definition}    & \textbf{Purpose} & \textbf{How to measure} & \textbf{Metric}\\
        \midrule

        $\D$
        & \hpqa
        & Measure state of multihop reasoning
        & Evaluate trained $M$ on $D$
        & $S^{\tau}(M, D)$ \\

        \midrule

        $\PRBD$
        & Probing dataset of $\D$
        & Measure how much disconnected reasoning $M$ does on $\tau$ test of $D$
        & Evaluate $M$ on $\PRBtest(\D)$ in zero-shot setting or with inoculation 
        & $\text{DiRe}^{\tau}(M, \D)$ [Equation \ref{eqn:original-model-cheatability}] \\

        \cmidrule{3-5}

        & 
        & Measure how much cheatable $\tau$ test of $\D$ is via disconnected reasoning
        & Train and evaluate a strong NLP model on $\PRBtest(\D)$
        & $\text{DiRe}^{\tau}(\D)$ \quad  [Equation \ref{eqn:original-data-cheatability}]  \\

        \midrule

        $\TD$
        & Transformed dataset of $\D$
        & Measure truer state of multi-hop reasoning, by reducing the amount of cheatability compared to $D$
        & Evaluate trained $M'$ on $\TD$
        & $S^{\tau+\textrm{suff}}(M', \TD)$ [Section \ref{subsec:transformation}] \\

        \midrule

        $\PRBTD$
        & Probing dataset of $\TD$
        & Measure how much cheatable $\tau+\textrm{suff}$ test of $\TD$ is via disconnected reasoning
        & Train and evaluate a strong NLP model on $\PRBtestsuff(\TD)$
        & $\text{DiRe}^{\tau+\textrm{suff}}(\TD)$ [Equation \ref{eqn:transformed-data-cheatability}]\\

        \midrule
    \end{tabular}
    \caption{Summary of dataset variations we create, their purposes and how we use them. $\tau$ can be any test, but our experiments are with \ans + \sponly. $M$ and $M'$ are models that can take $\tau$ and $\tau+\textrm{suff}$ tests respectively. In our experiments, they are trained on $D$ and $T(D)$ respectively with supervision for $\tau$ and $\tau+\textrm{suff}$ respectively.
    }
    \label{table:framework-summary}
    \vspace{-10pt}
\end{table*}

\section{Experiments}
\label{sec:experiments}

To obtain a more realistic picture of the progress in \mfr, we compare the performance of the original Glove-based baseline model~\cite{hotpotqa} and a state-of-the-art transformer-based LM, XLNet~\cite{yang2019xlnetga} on the multi-hop QA dataset \hpqa~\cite{hotpotqa}. While it may appear that the newer models are more capable of \mfr (based on answer and support prediction tasks), we show most of these gains are from better exploitation of \discreas. Our proposed transformation reduces \discreas exploitable by these models and gives a more accurate picture of the state of \mfr. \added{To support these claims, we use our proposed dataset probes, transformations, and metrics summarized in Table~\ref{table:framework-summary}.}

\textbf{Datasets $\D$ and $\TD$:}
\hpqa is a popular multi-hop QA dataset with about 113K questions which has  spurred many models~\cite{qfe,dfgn,sae,hgn}. We use the \emph{distractor} setting where each question has a \emph{set} of 10 input paragraphs, of which two were used to create the multifact question. Apart from the answer span, each question is annotated with these two supporting paragraphs and the supporting sentences within them. As described in Sec.~\ref{subsec:transformation}, we use these supporting paragraph annotations as $F_s$ to create a transformed dataset $\TD$.\footnote{We do not use the sentence-level annotations as we found them to be too noisy for the purposes of transformation.}

\textbf{Models:}
We evaluate two models:
    \textbf{(1) XLNet-Base:} Since \hpqa contexts are 10 paragraphs long, we use XLNet, a model that can handle contexts longer than 1024 tokens. We train XLNet-Base to predict the answer, supporting sentences, supporting paragraphs, and the sufficiency label (only on transformed datasets).
     As shown in Table~\ref{table:qa-results} \added{of Appendix~\ref{app:qa-results}}, our model is comparable to other models of similar sizes on the \hpqa dev set.
    \textbf{(2) Baseline:} We re-implement the baseline model from \hpqa. It has similar answer scores and much better support scores than the original implementation (details in Appendix~\ref{app:xlnet_details}).

\textbf{Metrics:} 
We report metrics for standard tests for \hpqa: answer span prediction (\ans), support identification (paragraph-level: \psp, sentence-level: \ssp), as well as \emph{joint} tests \ans+\psp and \ans+\ssp.
For each of these, we show F1 scores, but trends are similar for EM scores.\footnote{See Appendix~\ref{app:em_numbers} for these metrics for all our results.} These metrics correspond to $\mtest(q, \cdot)$ in Section~\ref{sec:measuring} and to $S^{\tau}(M, D)$ in Table~\ref{table:framework-summary}. When evaluating on the probing or transformed datasets, we use the corresponding metrics shown in Table~\ref{table:framework-summary}.

\subsection*{Measuring Disconnected Reasoning}

We first use our \probe\ probe to estimate the amount of \discreas in \textbf{\hpqa models} \added{(Eqn.~\ref{eqn:original-model-cheatability})}.
For this, we train our models on $\D$ and evaluate them against $\PRBD$, the probe dataset, under three settings:
(1) zero-shot evaluation (no training on $\PRBD$),
(2) after fine-tuning on 1\% of $\PRBD$, and (3) after fine-tuning on 5\% of $\PRBD$. Since the model has never seen examples with the modified context used in the probe, the goal of fine-tuning or \emph{inoculation}~\cite{inoculation} is to allow the model to adapt to the new inputs, while not straying far from its original behavior on $\D$.

\begin{figure}[t]
    \centering
	\includegraphics[width=0.42\textwidth]{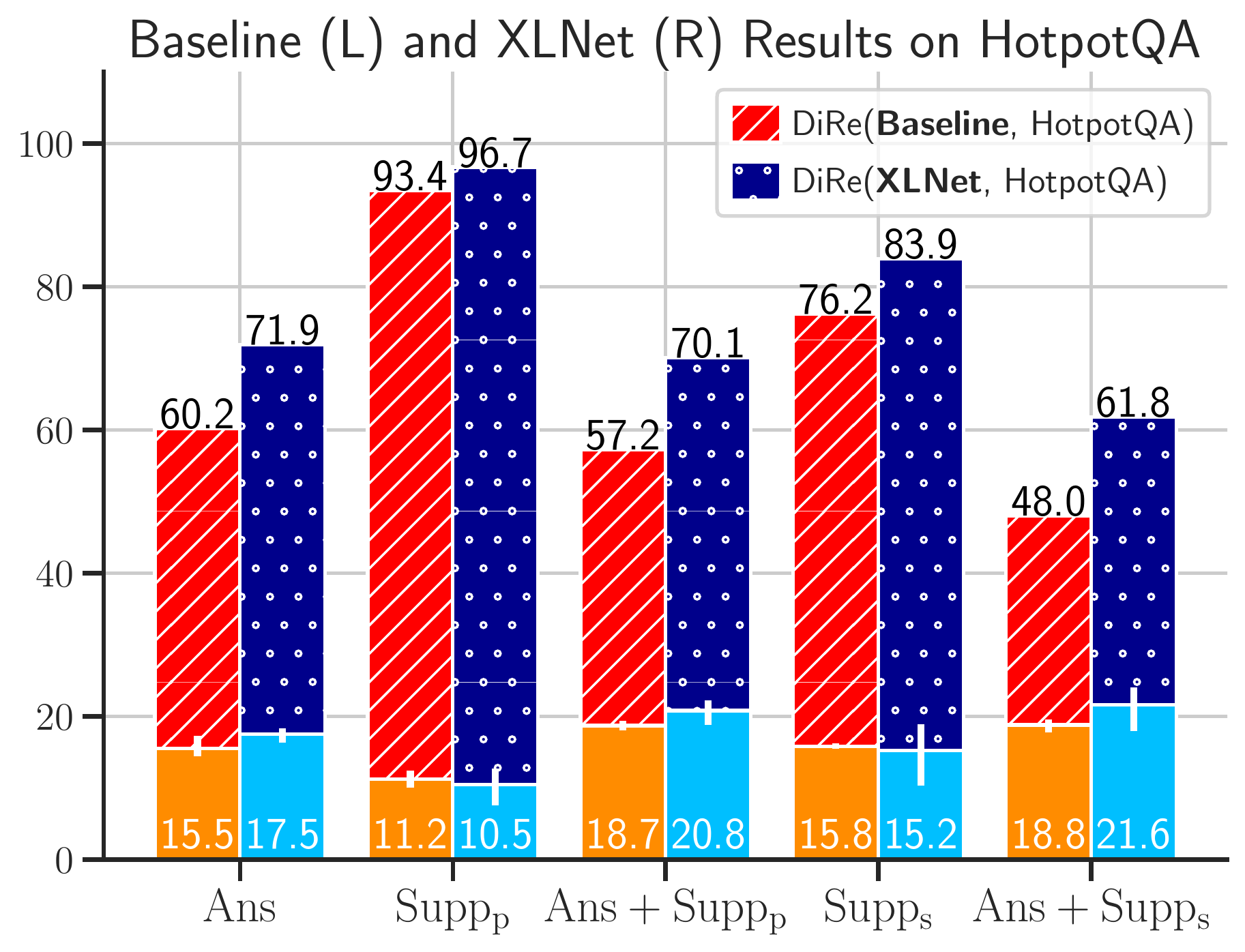}
	\caption{F1 scores for two models under various metrics. Progress on \hpqa from Baseline model to XLNet (entire bars) is largely due to progress in \discreas (upper, darker regions), with little change in \mfr (lower, ligher regions).
	}
	\label{fig:progress-due-to-disconnected-reasoning}
	\vspace{-10pt}
\end{figure}

Figure~\ref{fig:progress-due-to-disconnected-reasoning} summarizes the results. The \added{total heights of the} bars depict overall scores of the baseline and XLNet models on $\D$. The upper, darker regions depict the portion of the overall score achieved via disconnected reasoning as estimated by the \probe probe.\footnote{\added{This is a \emph{conditional} score as explained in Eqn.~(\ref{eqn:original-model-cheatability}).}
}
Their height is based on the average across the three fine-tuning settings, with white error margins depicting min/max. Importantly, results vary only marginally across the 3 settings. The lower, lighter regions show the remaining score, attributable to \mfr.

First, the amount of \mfr in XLNet is low---ranging from $10.5$ to $21.6$ F1
across the metrics. Second, even though the scores have improved going from the baseline model to XLNet, the amount of \mfr (lighter regions at the bottom) has barely improved. Notably, while the XLNet model improves on the \ans + \psp metric by 14 pts, the amount of \mfr has only increased by 3 pts!  
While existing metrics would \added{suggest} substantial progress in \mfr for \hpqa, the \probe probe shows that this is likely not the case---empirical gains are mostly due to higher \discreas. 

As a sanity check, we also train a Single-Fact XLNet model (Appendix~\ref{app:sf-xlnet}) that only reasons over one paragraph at a time---a model incapable of \mfr. This model achieves nearly identical scores on $\D$ as $\PRBD$, demonstrating that our \probe\ probe captures the extent of \discreas performed by a model (see Appendix~\ref{app:sf-results}).

\added{Next, we use the \probe probe to estimate how cheatable is the \textbf{\hpqa dataset} via disconnected reasoning (Eqn.~\ref{eqn:original-data-cheatability}). For this, we train and evaluate the powerful XLNet model on $\PRBD$.\footnote{The use of even stronger models is left to future work.} While the answer prediction test is known to be cheatable, we find that even the supporting fact (paragraph/sentence) identification test is highly cheatable (up to 91.2 and 75.7 F1, resp.).
}

\subsection*{Reducing Disconnected Reasoning}

Our automatic transformation reduces \discreas bias in the dataset and gives a more realistic picture of the state of \mfr. We show this by comparing how much score can a strong model (XLNet) achieve using \discreas on the original dataset, \added{by training it on $\PRBD$ and computing $\text{DiRe}^\tau(D)$} \added{(Eqn.~\ref{eqn:original-data-cheatability})}, and on the transformed dataset, \added{by training it on $\PRBTD$ and computing $\text{DiRe}^{\tau+\text{suff}}(\TD)$} \added{(Equation \ref{eqn:transformed-data-cheatability})}. \added{Training the model allows it to learn the kind of disconnected reasoning needed to do well on these probes, thus providing an upper estimate of \added{the cheatability of $\D$ and $\TD$} via disconnected reasoning.}

\begin{figure}[t]
    \centering
	\includegraphics[width=0.42\textwidth]{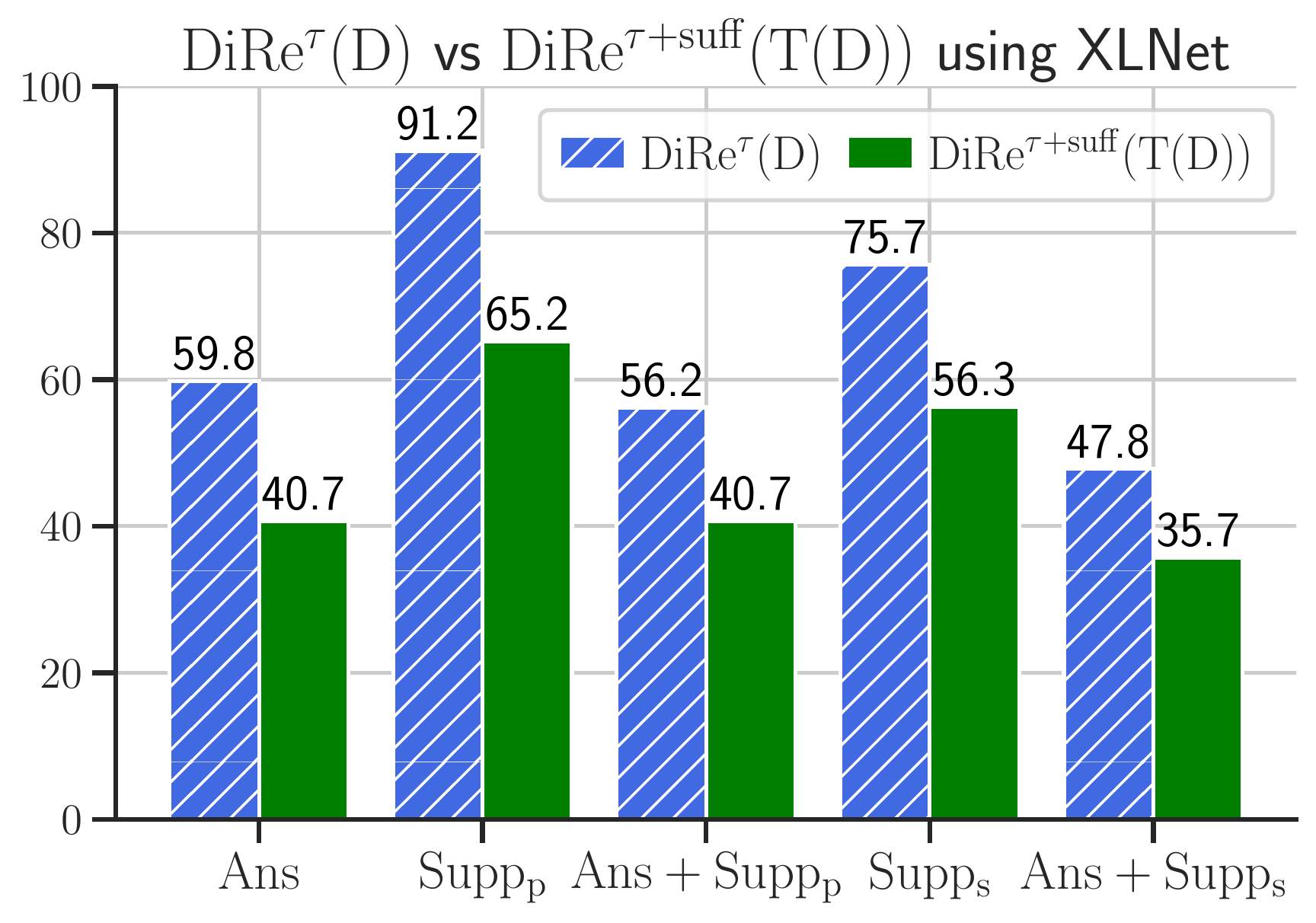}
	\caption{F1\added{-based \probe} scores of $\D$ and $\TD$ \added{using XLNet-Base}. Dataset transformation reduces disconnected reasoning bias, demonstrated by \probe scores being substantially lower on $\TD$ than on $\D$.}
	\label{fig:transformation-reduces-disconnected-bias}
	\vspace{-8pt}
\end{figure}

Figure ~\ref{fig:transformation-reduces-disconnected-bias} shows that the XLNet model's \probe\ score on the \ans+\psuff metric for $\TD$ is only 40.7, much lower compared to its \probe\ score of 59.8 on \ans for $\D$. 
 Across all metrics, $\TD$ is significantly less exploitable via \discreas than $\D$, drops ranging from 12 to 26 pts. 

\begin{figure}[t]
    \centering
	\includegraphics[width=0.38\textwidth]{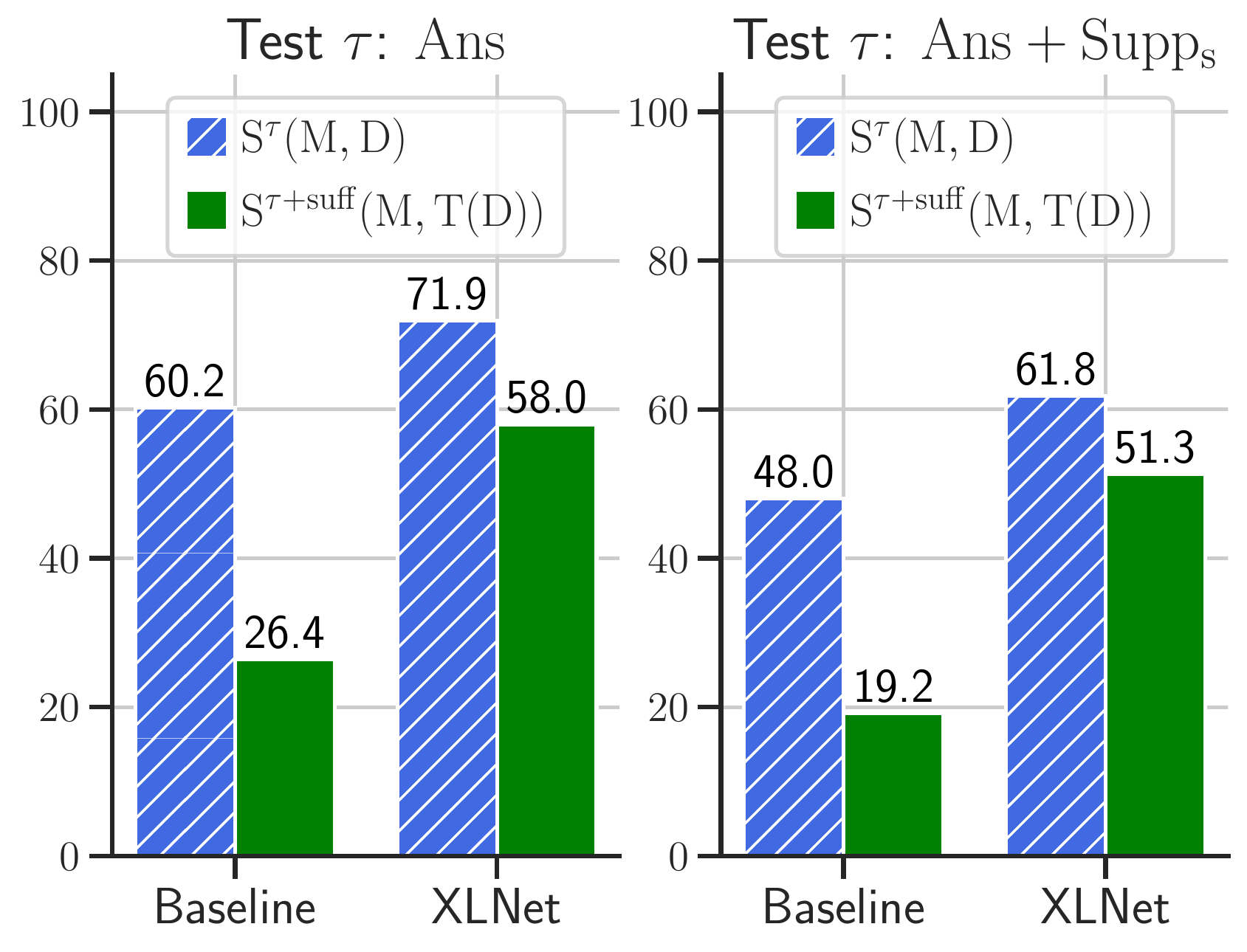}
	\caption{F1 scores of two models on $\D$ and $\TD$ under two common metrics. Transformed dataset is harder for both models since they rely on \discreas. The weaker, Baseline model drops more as it relies more heavily on \discreas.}
	\label{fig:transformed-is-harder}
	\vspace{-8pt}
\end{figure}

\subsection*{$\TD$ is a Harder Test of \MfR}

By reducing the amount of exploitable \discreas in $\TD$, we show that our transformed dataset is harder for models that have relied on \discreas.
Figure~\ref{fig:transformed-is-harder} shows that the transformed dataset is harder for both models across all metrics. Since a true-multihop model would naturally detect insufficient data, the drops in performance on $\TD$ show that the current model architectures \added{when trained on $\D$} are reliant on disconnected reasoning. The weaker baseline model has substantially lower scores on $\TD$, suggesting that simple models cannot get high scores. 

Single-Fact XLNet (the model incapable of \mfr as described earlier) also sees a big drop (-23 F1 pts on \ans) going from $\D$ to $\TD$ -- almost all of which was caught as \discreas by our \probe probe (see Appendix~\ref{app:sf-results}).

\subsubsection*{$\TD$ is Hard for the Right Reasons}
\label{ssec:right-for-right-reasons}

Our transformation makes two key changes to the original dataset $\D$: (C1) adds a new \emph{sufficiency test}, and (C2) uses a grouped metric over a set of \emph{contrastive examples}. We \added{argue} that these \added{changes by themselves} do not result in a score drop independent of the model's ability to perform \mfr \added{(details in Appendices~\ref{app:human-eval} and \ref{app:trivial-transformation})}.

\subsection*{Transformation vs.\ Adversarial Augmentation}

\begin{figure}[t]
    \centering
	\includegraphics[width=0.47\textwidth]{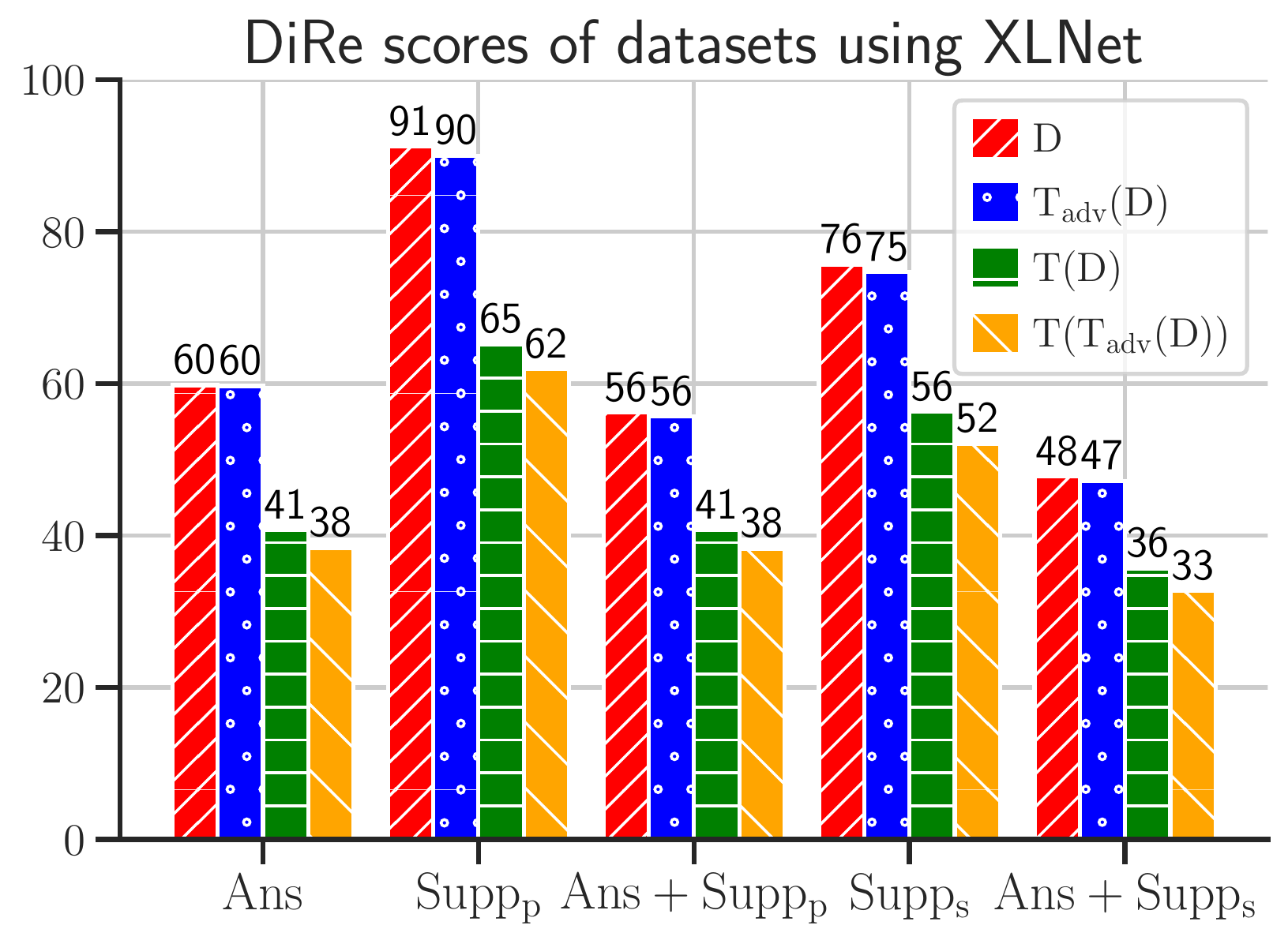}
	\caption{F1\added{-based \probe} score on various metrics \added{using XLNet-base} for $\D$, adversarial $\TadvD$, transformed $\TD$, and transformed adversarial $\TTadvD$. Transformation here is more effective than, and complementary to, adversarial augmentation.
	\label{fig:adversarial}
    }
	\vspace{-10pt}
\end{figure}

An alternate approach to reduce \discreas is via adversarial examples for single-fact models. \citet{jiang2019avoiding} proposed such an approach for \hpqa.
As shown in Figure~\ref{fig:adversarial}, our transformation results in a larger reduction in \discreas across all metrics; e.g., the XLNet model only achieves a \added{\probe} score \added{(metric: \ans+\ssp)} of 36 F1 on $\TD$ as compared to 47 F1 on $\Tadv(\D)$, \added{computed using Eqns.~(\ref{eqn:transformed-data-cheatability}) and (\ref{eqn:original-data-cheatability}), resp}. Moreover, since our approach can be applied to any dataset with supporting fact annotations, we can even transform the adversarial dataset, further reducing the \probe\ score to 33 F1. 

\section{Conclusions}

Progress in \added{\mh QA under the reading comprehension setting} relies on understanding and quantifying the types of undesirable reasoning current models may perform. This work introduced a formalization of disconnected reasoning, a form of bad reasoning prevalent in \mh models. It showed that a large portion of current progress in \mfr can be attributed to disconnected reasoning. Using a notion of contrastive sufficiency, it showed how to automatically transform existing support-annotated \mh datasets to create a more difficult and less cheatable dataset that results in reduced disconnected reasoning.

\added{Our probing and transformed dataset construction assumed that the context is an unordered set of facts. Extending it to a \emph{sequence} of facts (e.g., as in MultiRC~\cite{MultiRC2018}) requires accounting for the potential of new artifacts by, for instance, carefully replacing rather than dropping facts.  Additionally, for factual \added{reading comprehension} datasets where \added{the correct answer can be arrived at without consulting all annotated facts in the input context}, our probe will unfairly penalize a model that \added{uses implicitly known facts,} even if it correctly connects information across these facts. However, our transformation alleviates this issue: a model that connects information
will have an edge in determining the sufficiency of the given context.
We leave further exploration to future work.}

It is difficult to create large-scale multihop QA datasets that do not have unintended artifacts, and it is also difficult to design models that do not exploit such shortcuts. Our results suggest that carefully devising tests that probe for desirable aspects of \mfr is an effective way forward.

\subsection*{Acknowledgments}

\added{This work was supported in part by the National Science Foundation under grants IIS-1815358 and CCF-1918225. Computations on beaker.org were supported in part by credits from Google Cloud. We thank the anonymous reviewers and Greg Durrett for feedback on earlier versions of this paper.}

\bibliography{DiRe}
\bibliographystyle{acl_natbib}

\clearpage

\appendix
\section{Probe and Transformation Details}
\label{app:technical}

\begin{figure*}[t] 
    \small
    \begin{framed}
        \textbf{Original Dataset} $D$\\
        $\Rightarrow$ Question $q = (Q,C; A)$ in $D$ is assumed to be annotated with supporting facts $\{f_1, f_2\}$.
        \\
        
        \textbf{Probing Dataset} $\mathbb{P}_{\textrm{ans}+\textrm{supp}}(D)$ for Answer Prediction and Support Identification tests:\\
        $\Rightarrow$ Probing question collection $\mathbb{P}_{\textrm{ans}+\textrm{supp}}(q)$ has only one group, corresponding to the unique bi-partition $\{\{f_1\}, \{f_2\}\}$, containing:
        \begin{enumerate}[noitemsep]
            \item $(Q, C \setminus \{f_1\}; \Lansifpresent{=}A, \Lsupp{=}\{f_2\})$
            \item $(Q, C \setminus \{f_2\}; \Lansifpresent{=}A, \Lsupp{=}\{f_1\})$
        \end{enumerate}
        
        \textbf{Transformed Dataset} $\T(D)$ for evaluating Constrastive Support Sufficiency:\\
        $\Rightarrow$ Transformed question group $\mathbb{T}(q)$ in $\mathbb{T}(D)$ is defined using a single replacement fact $f_r \in C \setminus \{f_1, f_2\}$:
        \begin{enumerate}[noitemsep]
            \item $(Q, C \setminus \{f_r\};\, \Lans{=}A, \Lsupp{=}F_s, \Lsuff{=}1)$
            \item $    (Q, C \setminus \{f_1\};\, \Lsuff{=}0)$
            \item $(Q, C \setminus \{f_2\};\, \Lsuff{=}0)$
        \end{enumerate}

        \textbf{Probing Dataset} $\mathbb{P}_{\textrm{ans}+\textrm{supp}+\textrm{suff}}(\T(D))$ for all three tests:\\
        $\Rightarrow$ Probing question collection $\mathbb{P}_{\textrm{ans}+\textrm{supp}+\textrm{suff}}(\T(q))$ for the transformed question $\T(q)$ has only one group, corresponding to the unique bi-partition $\{\{f_1\}, \{f_2\}\}$, and is defined as:
        \begin{enumerate}[noitemsep]
            \item $(Q, C \setminus \{f_1, f_r\};\, \Lansifpresent{=}A, \Lsupp{=}\{f_2\}, \Lsuffprb{=}0)$
            \item $(Q, C \setminus \{f_2, f_r\};\, \Lansifpresent{=}A, \Lsupp{=}\{f_1\}, \Lsuffprb{=}0)$
            \item $(Q, C \setminus \{f_1, f_2\};\, \Lsuffprb{=}-1)$
        \end{enumerate}
    \end{framed}
    \caption{Proposed dataset transformation and probes for the case of $|F_s| = 2$ supporting facts.}
    \label{fig:2-fact-reasoning}
\end{figure*}

\begin{figure*}[t]
    \centering
	\includegraphics[width=0.8\textwidth]{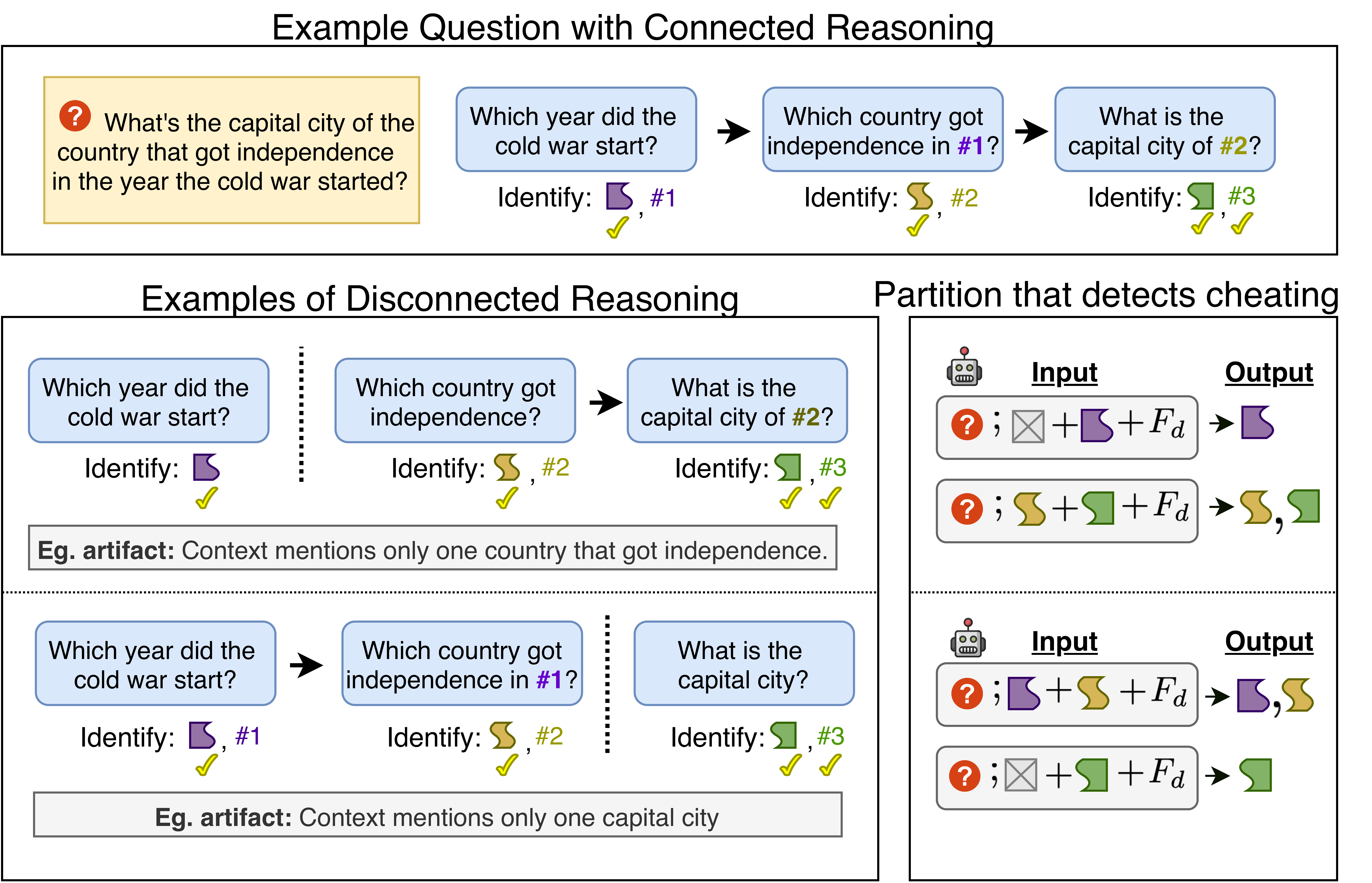}
	\caption{
	    Generalization of disconnected reasoning to a 3-fact reasoning question. As shown in the bottom half, a model could perform \mf reasoning on two disjoint partitions to answer this question. We consider such a model to be performing \discreas as it does not use the entire chain of reasoning and relies on artifacts (specifically, it uses 1-fact and 2-fact reasoning, but not 3-fact reasoning). For each of the two examples, there exists a fact bi-partition (shown on the right) that we can use to detect such reasoning as the model would continue to produce all the expected labels even under this partition.
	}
	\label{fig:dire-generalization}
\end{figure*}

\subsection{Probes and Transformation for $|F_s| = 2$}

Figure~\ref{fig:2-fact-reasoning} summarizes in a single place all probes and transformation discussed for the case of two supporting facts ($|F_s| = 2$).

\subsection{Need for Considering All Bi-partitions}
\label{app:all-bipartitions}

Figure~\ref{fig:dire-generalization} illustrates support bi-partitions and two examples of disconnected reasoning for a 3-hop reasoning question. It highlights the need for considering \emph{every} bi-partition in the \direcondition. For instance, if we only consider partitions that separate the purple (\purplefact) and yellow (\yellowfact) facts, then the model performing the lower example of disconnected reasoning would not be able to output the correct labels in any partition and would thus appear to not satisfy the \direcondition. We would therefore not be able to detect that it is doing \discreas.

\subsection{Transformation for $|F_s| \geq 2$}
\label{app:transform-general-k}

The Contrastive Support Sufficiency Transform described in Section~\ref{subsec:transformation} for the case of two supporting facts can be generalized as follows for $|F_s| \geq 2$. There are two differences. First, there are $2^{|F_s|} - 2$ choices of proper subsets of $F_s$ that can be removed to create insufficient context instances. Second, these subsets are of different sizes, potentially leading to unintended artifacts models can exploit. Hence, we use \emph{context length normalization} 
to ensure every context has precisely $|C| - |F_s| + 1$ facts. To this end, let $F_r$ be a fixed, uniformly sampled subset of $C \setminus F_s$ of size $|F_s| - 1$ that we will remove for the sufficient context instance.\footnote{We assume $|C| \geq 2 |F_s| - 1$.} Further, for each non-empty insufficient context $F_{s1} \subset F_s, F_{s1} \neq \phi$, let $F_{r1}$ denote a fixed, uniformly sampled subset of $F_r$ of size $|F_s| - |F_{s1}| - 1$. The transformed group $\T(q)$ contains the following $2^{|F_s|} - 1$ instances:
\begin{eqnarray}
    (Q, C \setminus F_r;\, \Lans{=}A, \Lsupp{=}F_s, \Lsuff{=}1)\\
    (Q, C \setminus (F_{s1} \cup F_{r1});\, \Lsuff{=}0) \text{\ \ \ for all\ } F_{s1}
\end{eqnarray}

Note that $|F_r| = |F_{s1}| + |F_{r1}| = |F_s| - 1$ by design, and therefore all instances have exactly $|C| - |F_s| + 1$ facts in their context.

Similar to the case of $|F_s| = 2$, for any performance metric 
\added{$\mtest(q, \cdot)$} of interest in $D$ (e.g., answer EM, support F1, etc.), the corresponding \emph{transformed metric} 
\added{$\mtestt(q, \cdot)$} operates in a conditional fashion: it equals $0$ if any $\Lsuff$ label in the group is predicted incorrectly, and equals 
\added{$\mtest(q_\textrm{suff}, \cdot)$} otherwise, where $q_\textrm{suff}$ denotes the unique sufficient context instance in $\T(q)$.

\subsection{Probing $\T(D)$ for $|F_s| = 2$}
\label{app:probe-T}

A model $M$ meets the \direcondition for \CSSTAbb when given an input context $C'$, it can correctly predict whether: (i) $C'$ contains $F_{s1}$, even when $F_{s2}$ is not in $C'$; (ii) $C'$ contains $F_{s2}$, even when $F_{s1}$ is not in $C'$; and (iii) $C'$ contains neither $F_{s1}$ nor $F_{s2}$.
Intuitively, if $M$ can do this correctly, then it has the information needed to correctly identify support sufficiency for all instances in the transformed group $\T(q)$, without relying on interaction between $F_{s1}$ and $F_{s2}$.

This leads to the following probe for $\T(D)$, denoted $\PRBall(\T(D))$ (sometimes simply $\PRB(\T(D))$ for brevity) and described here for the case of $F_s = \{f_1, f_2\}$.\footnote{Appendix~\ref{app:probe-T-general-k} describes the probe for $|F_s| \geq 2$.} Let $f_r$ be the fact used for context length normalization in the transformed group $\T(q)$. 
Similar to Eqns.~(\ref{eqn:2fact-probe-1}) and (\ref{eqn:2fact-probe-2}) in Section~\ref{subsec:dire-probe}, the probing dataset contains a group $\PRB(\T(q))$ of instances corresponding to the unique bi-partition $\{\{f_1\}, \{f_2\}\}$ of $F_s$:
\begin{align*}
    (Q, C \setminus \{f_1, f_r\};\, & \Lansifpresent{=}A, \Lsupp{=}\{f_2\}, \Lsuffprb{=}0)\\
    (Q, C \setminus \{f_2, f_r\};\, & \Lansifpresent{=}A, \Lsupp{=}\{f_1\}, \Lsuffprb{=}0)\\
    (Q, C \setminus \{f_1, f_2\};\, & \Lsuffprb{=}-1)
\end{align*}
%
$\Lansifpresent{}$, as before, is an optional label that is included in the instance only if $A$ is present in the supporting facts retained in the context of that instance.

We use the notation $\Lsuffprb$ here to highlight that this label is semantically different from $\Lsuff$ in $\T(D)$, in the sense that when $\Lsuffprb = 0$, the model during this probe is expected to produce the partial support and the answer (if present in the context). When not even partial support is there, the output label is $\Lsuffprb = -1$ and we don't care what the model outputs as the answer or supporting facts. Note that the label semantics being different is not an issue, as the probing method involves training models on the probe dataset. 

The \emph{joint} grouped metric here considers the sufficiency label, along with any standard test(s) of interest (answer prediction, support identification, or both). Denoted 
\added{$\mtestpt(q, \cdot)$},
it is defined as follows: similar to the conditional nature of the transformed metric
\added{$\mtestt(q, \cdot)$}, a model receives a score of 0 on the above group if it predicts the $\Lsuffprb$ label incorrectly for any instance in the group. Otherwise, we consider only the partial support instances (those with $\Lsuffprb=0$) in the group, which we observe are identical to the un-transformed probe group $\PRBanssupp(q; \{f_1\})$ when ignoring the sufficiency label, and apply the grouped probe metric
\added{$\mtestp$} from Section~\ref{subsec:dire-probe} to this subset of instances.

\subsection{Probing $\T(D)$ for $|F_s| \geq 2$}
\label{app:probe-T-general-k}

The probe for \discreas in the transformed dataset $\T(D)$ described in 
Appendix~\ref{app:probe-T}
for the case of two supporting facts can be generalized as follows for $|F_s| \geq 2$. For each proper bi-partition $\{F_{s1}, F_{s2}\}$ of $F_s$, we consider two partial contexts, $C \setminus F_{s1}$ and $C \setminus F_{s2}$, and one where not even partial support is present, $C \setminus (F_{s1} \cup F_{s2})$.

Recall that when constructing $\T(D)$, we had associated non-supporting facts $F_{r1}$ and $F_{r2}$ (both chosen from $F_r$) with supporting facts $F_{s1}$ and $F_{s2}$, respectively, and had additionally removed them from the respective input contexts for length normalization. For the partial context instances in the probe, we choose another non-supporting fact $f_{r1} \in F_r \setminus \cup F_{r1}$, and combine it with $F_{r1}$ to obtain $F'_{r1} = F_{r1} \cup \{f_{r1}\}$; similarly define $F'_{r2}$.

For each $q \in D$, the probing dataset contains a collection $\PRB(\T(q))$ of $2^{|F_s|-1} - 1$ groups of instances, where each group corresponds to one proper bi-partition of $F_s$. For the bi-partition $\{F_{s1}, F_{s2}\}$, the group, denoted $\PRB(\T(q); F_{s1})$, contains the following instances, each of which has exactly $|C| - |F_s|$ facts in its context:

\begin{enumerate}[noitemsep]
    \item $(Q, C \setminus (F_{s1} \cup F'_{r1});\,$\\
        \hspace*{3ex} $\Lansifpresent{=}A, \Lsupp{=}F_{s2}, \Lsuffprb{=}0)$
    \item $(Q, C \setminus (F_{s2} \cup F'_{r2});\,$\\
        \hspace*{3ex} $\Lansifpresent{=}A, \Lsupp{=}F_{s1}, \Lsuffprb{=}0)$
    \item $(Q, C \setminus F_s;\, \Lsuffprb{=}-1)$
\end{enumerate}
The semantics of $\Lansifpresent{}$ and $\Lsuffprb{}$ remain the same as for the case of $|F_s| = 2$.

The grouped metric for this bi-partition, denoted
\added{$\mtestpt(q, \prediction(\PRB(\T(q); F_{s1})))$}, captures whether the model exhibits correct behavior on the entire group (as discussed for the case of $|F_s| = 2$). The overall probe metric, 
\added{$\mtestpt(q, \cdot)$}, continues to follow Eqn.~(\ref{eq:probe-metric}) and captures the disjunction of undesirable behavior across all bi-partitions.

\section{XLNet QA Model Details}
\label{app:xlnet_details}

\subsection{XLNet-Base (Full)} 

We concatenate all 10 paragraphs together into one long context with special paragraph marker token \texttt{[PP]} at the beginning of each paragraph and special sentence marker token at the beginning of each sentence in the paragraph. Lastly, the question is concatenated at the end of this long context. Apart of questions that have answer as a span in the context, \hpqa also has comparison questions for which the answer is "yes" or "no" and it's not contained in the context. So we also prepend text \texttt{"<yes> <no>"} to the context to deal with both types of questions directly by answer span extraction. Concretely, we have, \texttt{[CLS] <yes> <no> [PP] [SS] sent1,1 [SS] sent1,2 [PP] [SS] sent2,1 [QQ] q}. 

We generate logits for each paragraph and sentence by passing marker tokens through feedforward network. Supporting paragraphs and sentences are supervised with binary cross entropy loss. Answer span extraction is using standard way \cite{devlin-etal-2019-bert} where span start and span end logits are generated with feedforward on each token and it's supervised with cross entropy loss. We use first answer occurrence among of the answer text among the supporting paragraphs as the correct span. This setting is very similar to recent work \cite{Beltagy2020Longformer}, and our results in Table~\ref{table:qa-results}, show that this model achieves comparable accuracy to other models with similar model complexity. We haven't done any hyperparameter (learning rate, num epoch) tuning on the development set because of the expensive runs, which could explain the minor difference.

For predicting sufficiency classification, we use feedforward on \texttt{[CLS]} token and train it with cross entropy loss. In our transformed dataset, because \hpqa has K=2, there are twice the number of instances with insufficient supporting information than the instances with insufficient supporting information. So during \textit{training} we balance the number of insufficient instances by dropping half of them. 

\subsection{XLNet-Base (Single Fact)}
\label{app:sf-xlnet}
To verify the validity of our tests, we also evaluate a variant of XLNet \emph{incapable of Multifact reasoning}. Specifically, we train our XLNet model that makes predictions one paragraph at a time (similar to ~\citet{min2019compositional}). Although these previous works showed that answer prediction is hackable, we adapt it to predict supporting facts and sufficiency as well.

Specifically, we process the following through the XLNet transformer \texttt{[CLS] <yes> <no> [PP] [SS] sent1,1 [SS] sent1,2 [QQ] q} for each paragraph. We then supervise \texttt{[PP]} tokens for two tasks: identify if paragraph is a supporting paragraph and identify if paragraph has the answer span (for yes/no question both supporting paragraphs are supervised to be having the answer). We then select top ranked paragraph for having the answer and generate the best answer span. Similarly, select top two ranked paragraphs for having being supporting and predict the corresponding supporting sentences. The logits for answer span and supporting sentences are ignored when the paragraph doesn't have the answer and is not supporting respectively. We train for three losses jointly: (i) ranking answer containing paragraph, (ii) ranking supporting paragraphs (iii) predicting answer from answer containing paragraph (iv) predicting supporting sentences from supporting paragraphs. We use binary cross entropy for ranking of paragraphs, so there's absolutely no interaction the paragraphs in this model. To get the sufficiency label, we apply check if the sufficiency classification label based on the number of supporting paragraphs predicted\footnote{This heuristic exploits the fixed number of hops=2 and doesn't need any training on the sufficiency label. We use this heuristic because we want to predict sufficiency label without interaction across \textit{any} of the facts.}. For original dataset, if $\mid$ predicted(\psp)$\mid$ $>$ 1, then $C=1$ otherwise $C=0$. For probing dataset, if $\mid$ predicted(\psp)$\mid$ $>$ 0, then $C=0$ otherwise $C=-1$.

\subsection{Glove-based Baseline}

We have re-implemented the baseline described in ~\cite{hotpotqa} in AllenNLP~\cite{Gardner2017AllenNLP} library. Unlike original implementation, which uses only answer and sentence support identification supervision, we also using paragraph supervision identification supervision. Additionally, we use explicit paragraph and sentence marker tokens as in our XLNet-based implementation, and supervise model to predict paragraph and sentences support logits via feedforward on these token marker representations. We train answer span identification by cross-entropy loss and both paragraph and sentence support identification with binary cross-entropy loss.


\subsection{QA Model Results}
\label{app:qa-results}

Table~\ref{table:qa-results} shows results for QA models. Our XLNet model is comparable to other models of similar sizes on the \hpqa dev set. Our implementation of RNN baseline model has answer scores similar to the reported ones, and has much better support identification scores than the original implementation.

\begin{table}[t]
    \centering
    \small
    \setlength\tabcolsep{3pt}
    \begin{tabular}{lccc}
        Model       &  Ans F1    & \ssp F1 & Joint F1 \\
        \midrule
        Baseline (reported)     & 58.3       & 66.7   & 40.9 \\
        QFE (BERT-Base)             & 68.7       & 84.7   & 60.6 \\
        DFGN (BERT-Base)            & 69.3       & 82.2   & 59.9 \\
        RoBERTa-Base     & 73.5       & 83.4   & 63.5 \\
        LongFormer-Base  & 74.3       & 84.4   & 64.4 \\
        \midrule
        Baseline (our)       & 60.2       & 76.2   & 48.0 \\
        XLNet-Base   & 71.9       & 83.9   & 61.8 \\
    \end{tabular}
    \caption{Performance of XLNet-Base compared to other transformer models (of similar size) on \hpqa. Our model scores higher than BERT-Base models QFE~\cite{qfe} and DFGN~\cite{dfgn}, and performs comparable to recent models using RoBERTa and Longformer~\cite{Beltagy2020Longformer}.
    }
    \label{table:qa-results}
\end{table}

\section{Implementation and Model Training}

All our models are implemented using AllenNLP~\cite{Gardner2017AllenNLP} library. For XLNet-base, we have also used Huggingface Transformers~\cite{Wolf2019HuggingFacesTS}. For all XLNet-base experiments, we train for two epochs, checkpointing every 15K instances and early stopping after 3 checkpoints of no validation metric improvement. For Glove-based baseline model, we do the same but for 3 epochs. For both models, effective batch size were 32. For XLNet-based model, we used learning rate of 0.00005 and linear decay without any warmup. The hyper-parameters were chosen as the default parameters used by hugging-face transformers to reproduce BERT results on SQuAD dataset. Our experiments were done using V100 gpus from Google Cloud. On average XLNet training runs took 2 days on 1 gpu and baseline model took less than 1 day.

\begin{figure*}[t]
    \centering
	\includegraphics[width=1.0\textwidth]{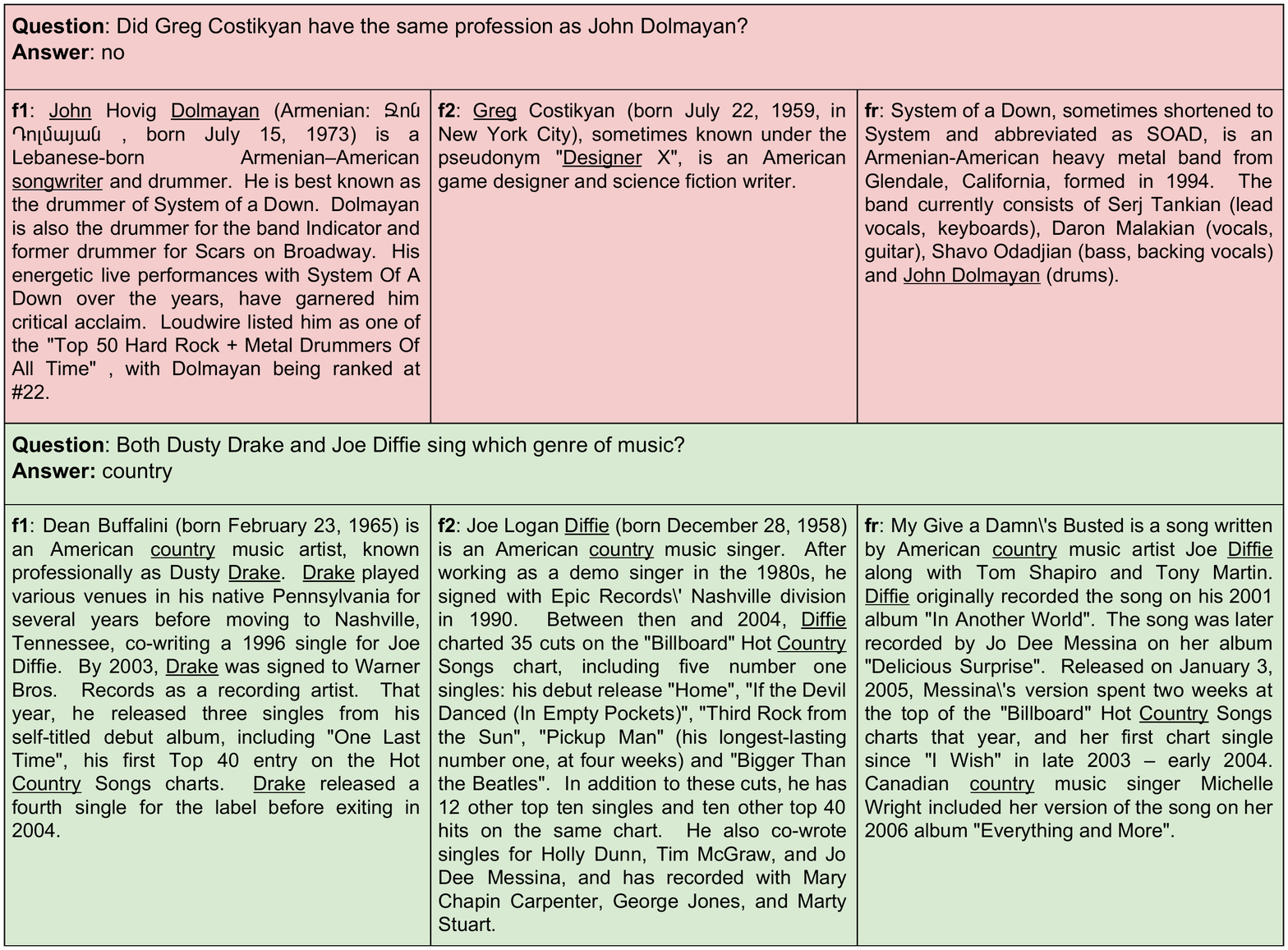}
	\caption{Two examples where we found a non-supporting fact provides an alternative support for answering the question. \textbf{f1} and \textbf{f2} are annotated supporting facts, but \textbf{fr} in C and \textbf{f1} form alternative support.}
	\label{fig:duplicates-example}
\end{figure*}

\section{Human Evaluation of Sufficiency Prediction}
\label{app:human-eval}

\added{The sufficiency test can cause a spurious drop if sufficiency labels are incorrect, i.e., the context is sufficient even after $f_1$ or $f_2$ is removed. To rule this out, we randomly evaluated (using MTurk) 115 paragraphs from $C \setminus F_s$, and found only 2 (1.7\%) could be used in place of $f_1$ or $f_2$ to answer the question. As we show below, this would result in only a marginal score drop compared to the roughly 20\% observed drops.}

\added{To estimate this,} we setup an annotation task on MTurk for turkers to annotate whether a pair of facts has sufficient information to arrive at an answer. For each question, we create three pair $(f_1, f_2)$, $(f_1, f_r)$ and $(f_r, f_2)$, where $f_1$ and $f_2$ are annotated supporting paragraphs, and $f_r$ is a randomly sample from total non-supporting paragraphs. The questions were taken from \hpqa development set. If for a question, annotators agree that both $(f_1, f_2)$ and $(f_1, f_r)$ are sufficient, we assume $f_r$ provides proxy (duplicate) information for $f_2$. Likewise for $(f_1, f_2)$ and $(f_r, f_2)$. 

Out of 115 examples questions (with annotator agreement), we found only 2 (1.7\%) of them to have a proxy fact in $f_r$. Figure ~\ref{fig:duplicates-example} shows these 2 examples. This shows that such proxy information is very rare in \hpqa. We next estimate the impact of these duplicates on the human score.

\subsection{Human Score Estimate on $\TD$}
Given the number of observed duplicates, we can now estimate the expected drop in human performance. For simplicity, lets consider only the Exact Match score where the human would get one point if they predict all the facts exactly. There are two scenarios where the sufficiency test in our transformed dataset would introduce more noise resulting in a drop in human score. 
\begin{enumerate}
    \item The original context was not actually sufficient: In this case the sufficiency label, $L_{\text{suff}} = 1$ would be incorrect and the human score on the sufficiency test for this example would be zero. However, in such a case, the human score on paragraph and sentence identification would also be zero. As a result, there would be no drop in human score relative to the original task
    \item The constrastive examples are actually sufficient: Due to a potential proxies of $f_1$ in $C \setminus{f_1}$, it is possible that our contrastive examples would be considered sufficient. While this would also effect the original dataset, its impact would be more extreme on our test. We focus on this scenario in more detail next. 
\end{enumerate}

Lets assume that there are $k$ such proxy paragraphs in any given context. In such a case, there is a $1/(k+1)$ chance that a human would select the annotated support paragraphs instead of these proxy paragraphs. So there is a $1/(k+1)$ chance that they get one point on the original task, but they would always get 0 points on our transformation. 

Given that we observed a proxy paragraph in 1.7\% of our annotated paragraphs, we can model the likelihood of observing $k$ proxy paragraphs with a binomial distribution. Specifically, since there are 8 distractor paragraphs in \hpqa, the probability of observing $k$ proxy paragraphs:
\[
  P(k) = \binom{8}{k} \times (0.017)^k \times (1- 0.017)^{8-k}
\]
So the expected drop in score would be given by:
\[
   \sum_{k=1}^8 P(k) \times \frac{1}{k+1} = 0.0628
\]
So the expected drop in human score is only 6.28\% whereas we observed about 18\% drop in EM scores as shown in Appendix~\ref{app:em_numbers}.

\section{XLNet (Single-Fact) Results}
\label{app:sf-results}
\begin{figure}[t]
    \centering
	\includegraphics[width=0.45\textwidth]{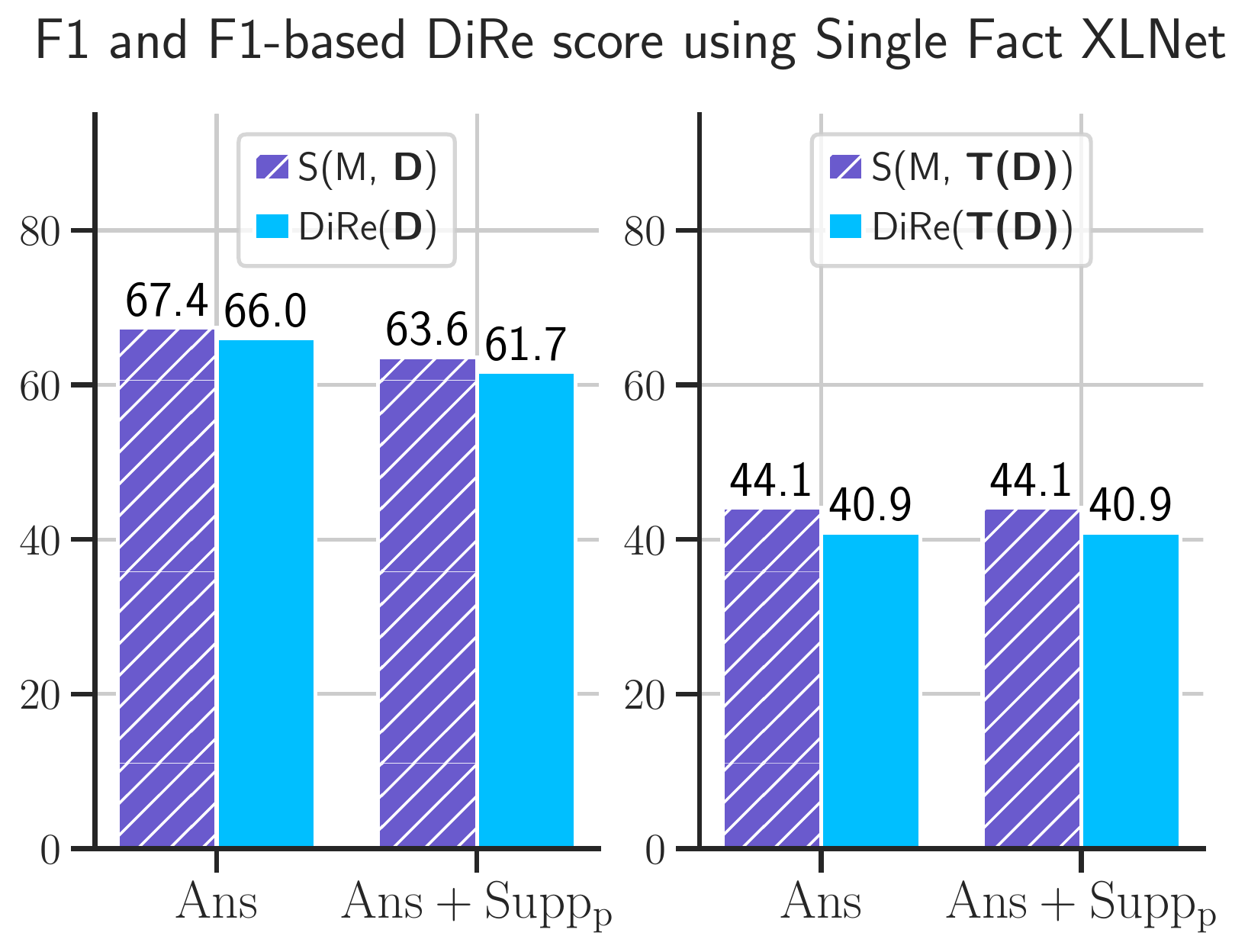}
	\caption{\added{F1 and F1-based \probe scores of $\D$ and $\TD$ using Single-Fact XLNet-base.}}
	\label{fig:single-fact}
\end{figure}

Figure~\ref{fig:single-fact} shows the results of our Single-Fact model on the original dataset $\D$ and the transformation $\TD$. On both the metrics, we can see that our \direprobe gets the almost the same score as the Single-Fact model, i.e., our probe can detect the \discreas bias in the Single-Fact model. Additionally, we can see that score of this Single-Fact model drops from 67.4 to 44.1, a drop of 23 F1 pts, going from $\D$ to $\TD$ (on the \ans metric).  This shows that our transformed dataset is less exploitable by a \discreas model. 

\section{Exact Match Numbers}
\label{app:em_numbers}
\begin{figure}[t]
    \centering
	\includegraphics[width=0.45\textwidth]{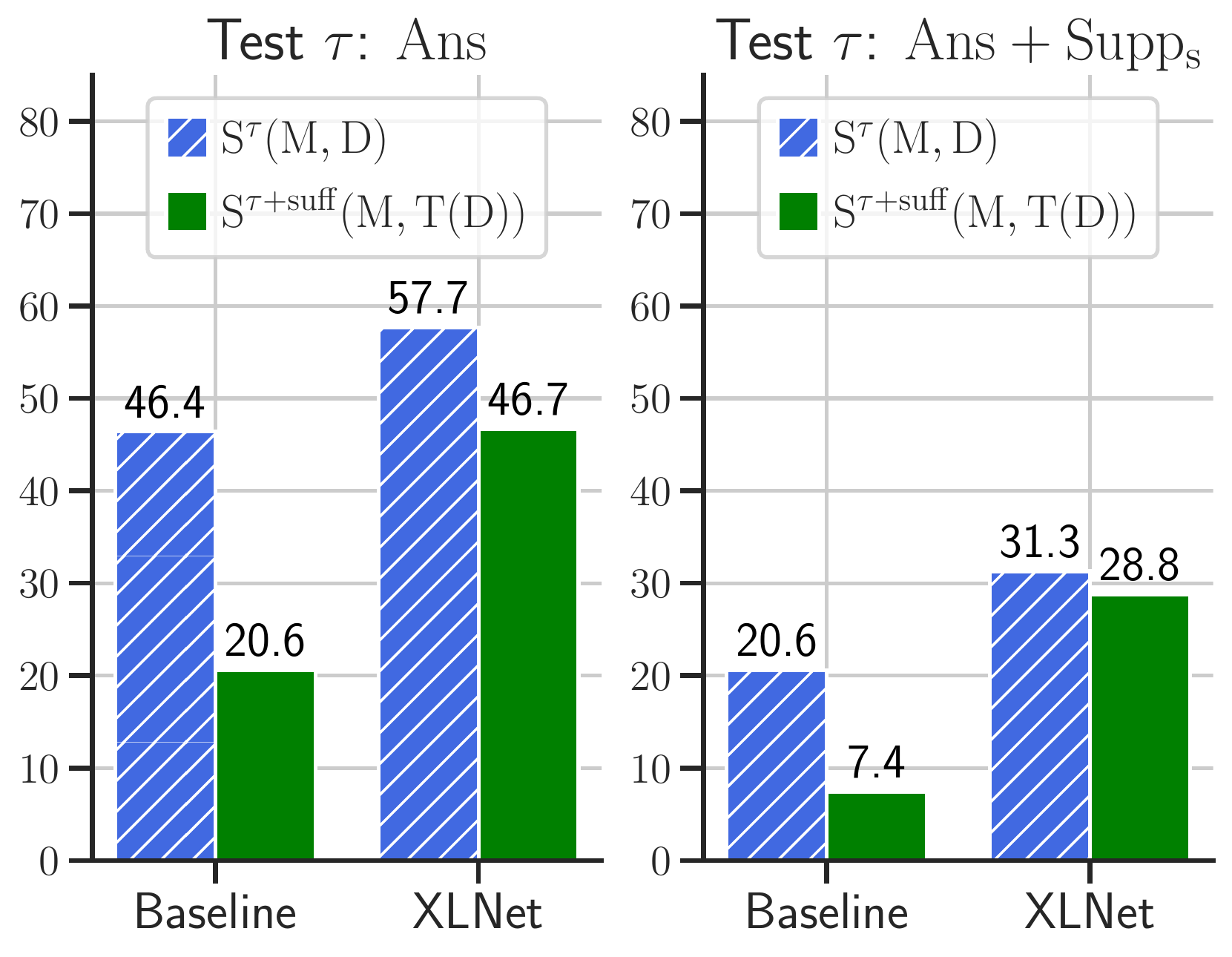}
	\caption{EM scores of two models on $\D$ and $\TD$ under two common metrics. Transformed dataset is harder for both models since they rely on \discreas. The weaker, Baseline model drops more as it relies more heavily on \discreas.}
	\label{fig:em-transformed-is-harder}
\end{figure}

Figure ~\ref{fig:em-transformed-is-harder} shows the EM scores of our models on the original dataset $\D$ and transformed dataset $\TD$. Consistent with our F1 metric, we can see large drops in model score going from $\D$ to $\TD$, showing that the transformation is harder for these models 

\begin{figure}[t]
    \centering
	\includegraphics[width=0.45\textwidth]{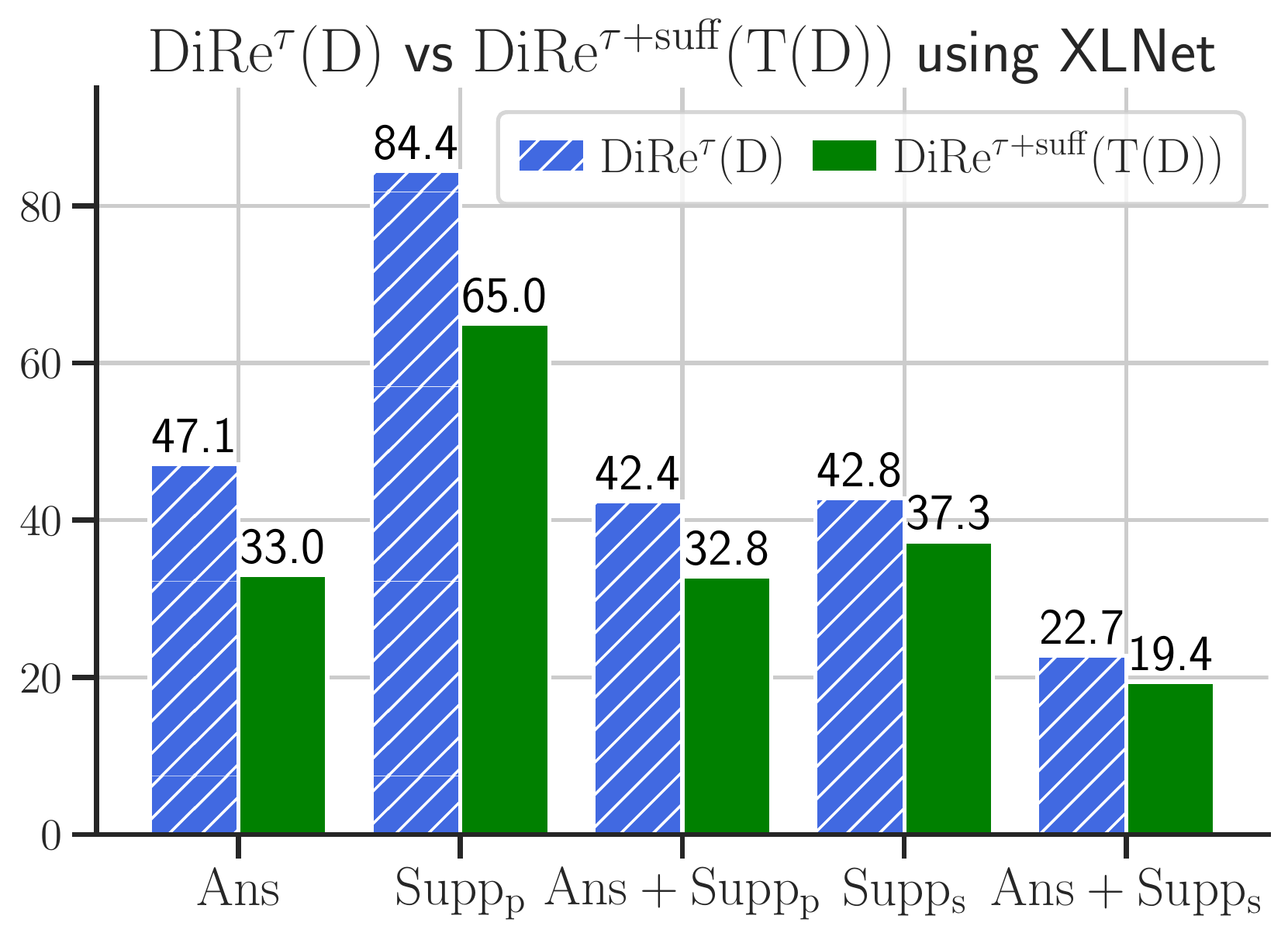}
	\caption{EM\added{-based \probe} scores of $\D$ and $\TD$ using XLNet-Base. Dataset transformation reduces disconnected reasoning bias, demonstrated by \probe scores being substantially lower on $\TD$ than on $\D$.}
	\label{fig:em-transformation-reduces-disconnected-bias}
\end{figure}

Figure ~\ref{fig:em-transformation-reduces-disconnected-bias} shows the \discreas bias in the XLNet-Base model trained on $\D$ and $\TD$ using the EM scores. Again, we see the same trend here -- the transformed dataset has a reduced \probe score indicating lower \discreas bias. 

\begin{figure}[t]
    \centering
	\includegraphics[width=0.45\textwidth]{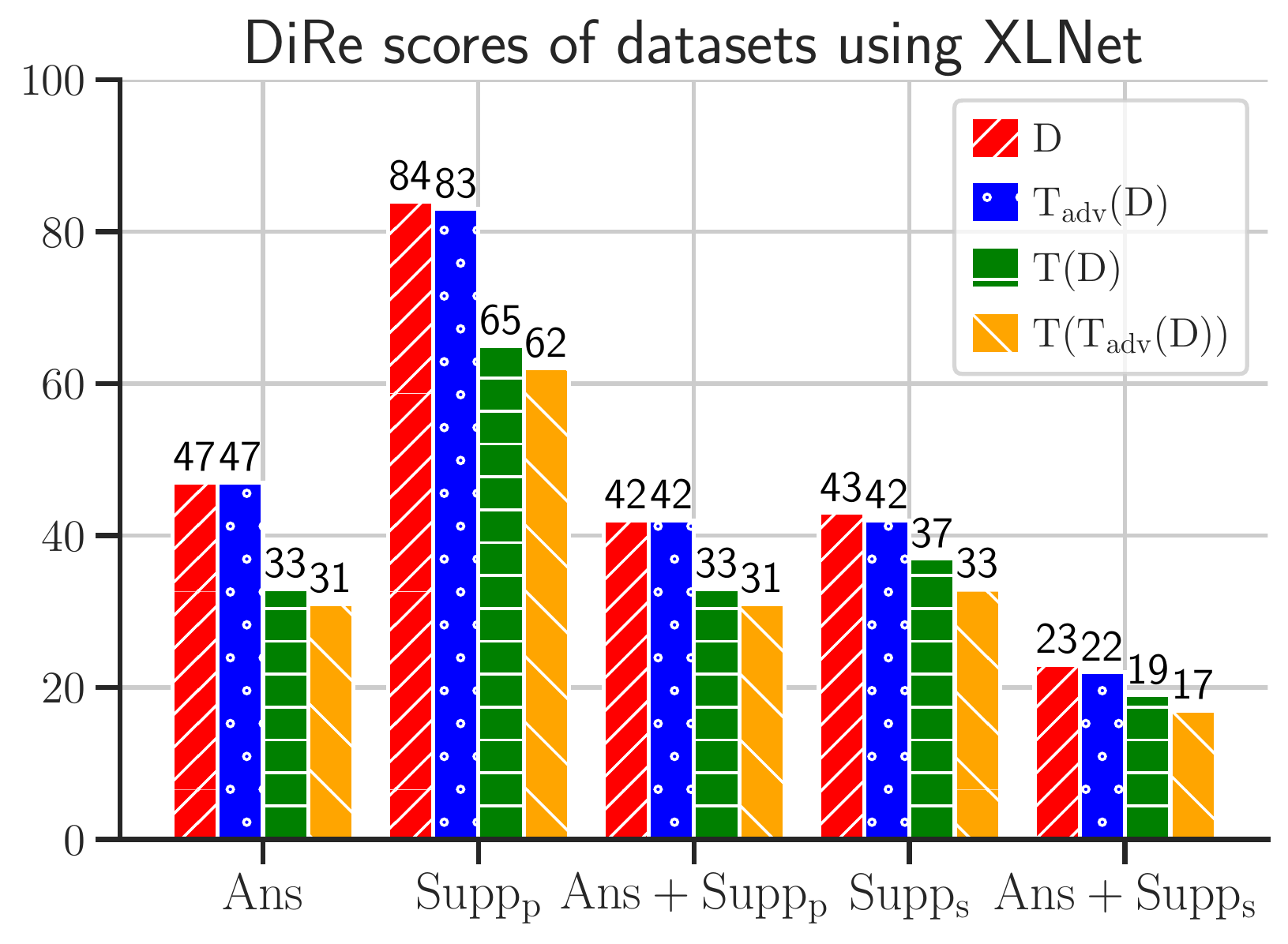}
	\caption{EM\added{-based \probe} score on various metrics \added{using XLNet-base} for four datasets: original $\D$, adversarial $\TadvD$, transformed $\TD$, and transformed adversarial $\TTadvD$. Transformation is more effective than, and complementary to, Adversarial Augmentation for reducing \probe scores.
	\label{fig:em-adversarial}
    }
\end{figure}

Finally, Figure~\ref{fig:em-adversarial} shows the impact of adversarial examples and the transformation on the EM scores. While the drops are lower due to the strictness of the EM scores, the trends are still the same -- adversarial examples have a minor impact on the \probe\ scores but transformation of the original dataset as well transformation of the adversarial examples results in a big drop in the \discreas bias.

\section{Grouped Metric on Trivial Transformation}
\label{app:trivial-transformation}

The grouped metric combines decisions over a set of instances and, one can argue, is therefore inherently harder. 
However, one can show that unless the instances within a group test for qualitatively different information, the grouped metric will not be necessarily lower than the single instance metric.

To support this claim, we compute grouped metric over a trivial transform that is similar to $\TD$ but does not involve the contrastive sufficiency prediction test. This trivial transform, denoted $\T_{trv}$, creates 3 copies of each instance but drops at random one non-supporting fact from each instance. Similar to $\TD$ in which we require the model to produce correct sufficiency labels for all 3 instances, here we require the model to produce correct answer and support on all 3 copies.\footnote{Note that our transformed dataset does not even require the answer and support labels on all the examples, making $\TD$, in some ways, easier than this dataset.}

\begin{enumerate}[noitemsep]
    \item $(Q, C \setminus \{f_{r_1}\};\, \Lans{=}A, \Lsupp{=}F_s)$
    \item $    (Q, C \setminus \{f_{r_2}\};\, \Lans{=}A, \Lsupp{=}F_s)$
    \item $(Q, C \setminus \{f_{r_3}\};\, \Lans{=}A, \Lsupp{=}F_s)$
\end{enumerate}

\begin{figure}[t]
    \centering
	\includegraphics[width=0.45\textwidth]{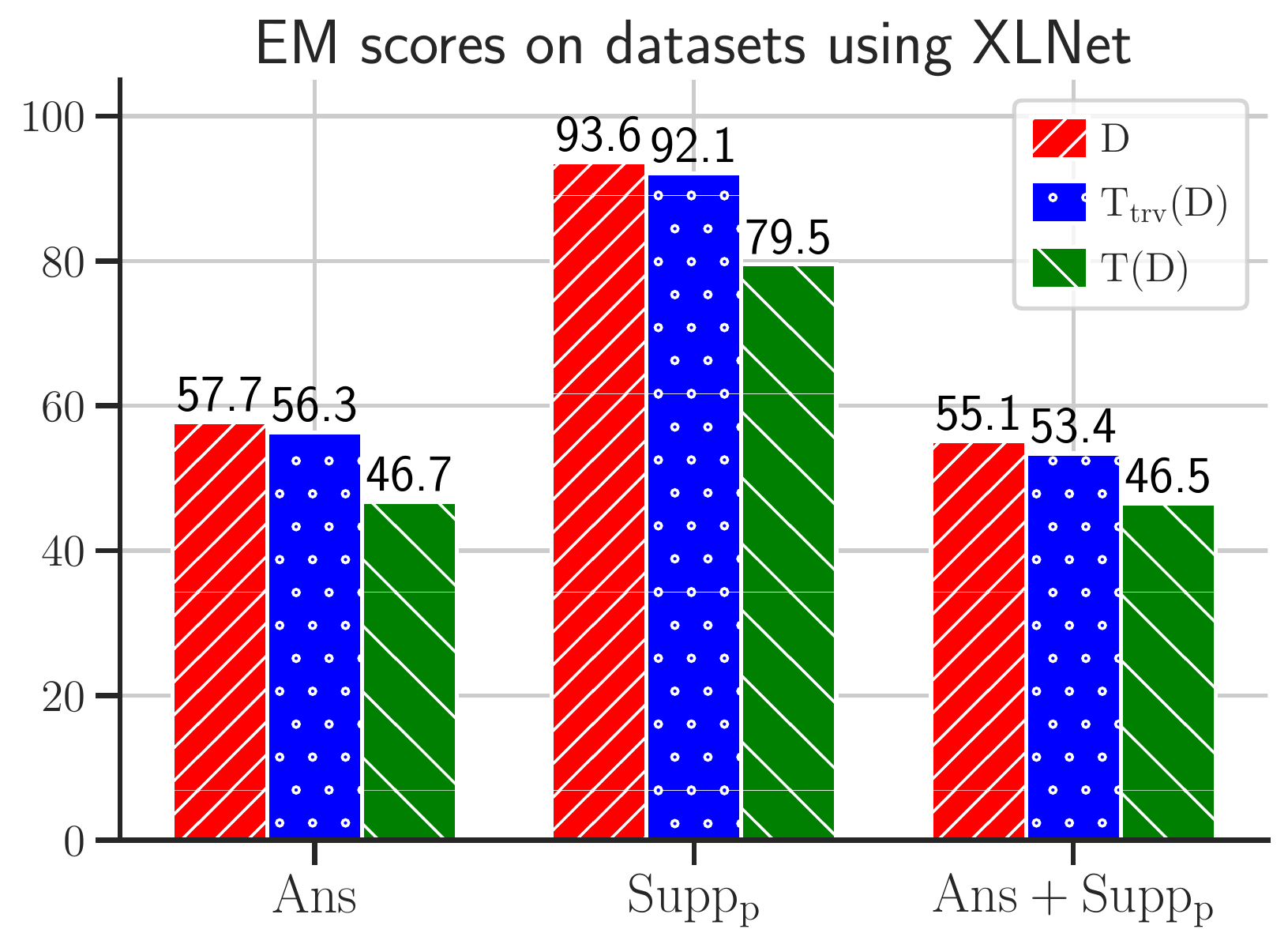}
	\caption{\added{EM scores of XLNet-base on three metrics for original \hpqa ($\D$), trivially transformed \hpqa ($\T_{trv}(\D)$) and our transformed \hpqa ($\TD$). Model scores barely drop from $\D$ to $\T_{trv}(\D)$ but significantly drop $\D$ to $\TD$ showing that drop of scores in $\TD$ is not simply a result of using a grouped metric}
	\label{fig:trivial-transformation}
    }
\end{figure}

In Figure ~\ref{fig:trivial-transformation}, we show the EM results corresponding to the respective grouped metrics for $\T_{trv}(\D)$ and $\TD)$. We see barely any drop of results from $\D$ to $\T_{trv}(\D)$, but do see significant drop going from $\D$ to $\TD$. This shows that adding a grouped metric over an arbitrary set of decisions would not make $\TD$ harder. \added{Hence, the drop in scores from $\D$ to $\TD$ is not simply a result of using a grouped metric}.

\end{document}